\documentclass{bmvc_sty_final/bmvc2k}

\title{It's all Relative: Monocular 3D Human Pose Estimation from Weakly Supervised Data}

\addauthor{Matteo Ruggero Ronchi*, Oisin Mac Aodha*, Robert Eng, Pietro Perona}{{mronchi,macaodha,reng,perona}@caltech.edu}{1}
\addauthor{California Institute of Technology}{}{1}

\runninghead{}{Monocular 3D Human Pose Estimation from Weakly Supervised Data}

\def\eg{\emph{e.g}\bmvaOneDot}

\def\ie{\emph{i.e}\bmvaOneDot}

\usepackage{amsfonts}

\begin{document}

\maketitle

%%%%%%%%% ABSTRACT
\vspace{-5mm}
\begin{abstract}
We address the problem of 3D human pose estimation from 2D input images using only weakly supervised training data.
Despite showing considerable success for 2D pose estimation, the application of supervised machine learning to 3D pose estimation in real world images is currently hampered by the lack of varied training images with corresponding 3D poses.  
Most existing 3D pose estimation algorithms train on data that has either been collected in carefully controlled studio settings or has been generated synthetically. 
Instead, we take a different approach, and propose a 3D human pose estimation algorithm that only requires relative estimates of depth at training time. 
Such training signal, although noisy, can be easily collected from crowd annotators, and is of sufficient quality for enabling successful training and evaluation of 3D pose algorithms.
Our results are competitive with fully supervised regression based approaches on the Human3.6M dataset, despite using significantly weaker training data.
Our proposed algorithm opens the door to using existing widespread 2D datasets for 3D pose estimation by allowing fine-tuning with noisy relative constraints, resulting in more accurate 3D poses.
\end{abstract}

%%%%%%%%% INTRODUCTION
\section{Introduction}\label{sec:intro}
\begin{figure*}[t]
    \centering
    \subfigure[Lifting 3D Pose Estimation]{  
        \centering
        \includegraphics[trim={0px 0px 0px 0px},clip,height=.18\textwidth]{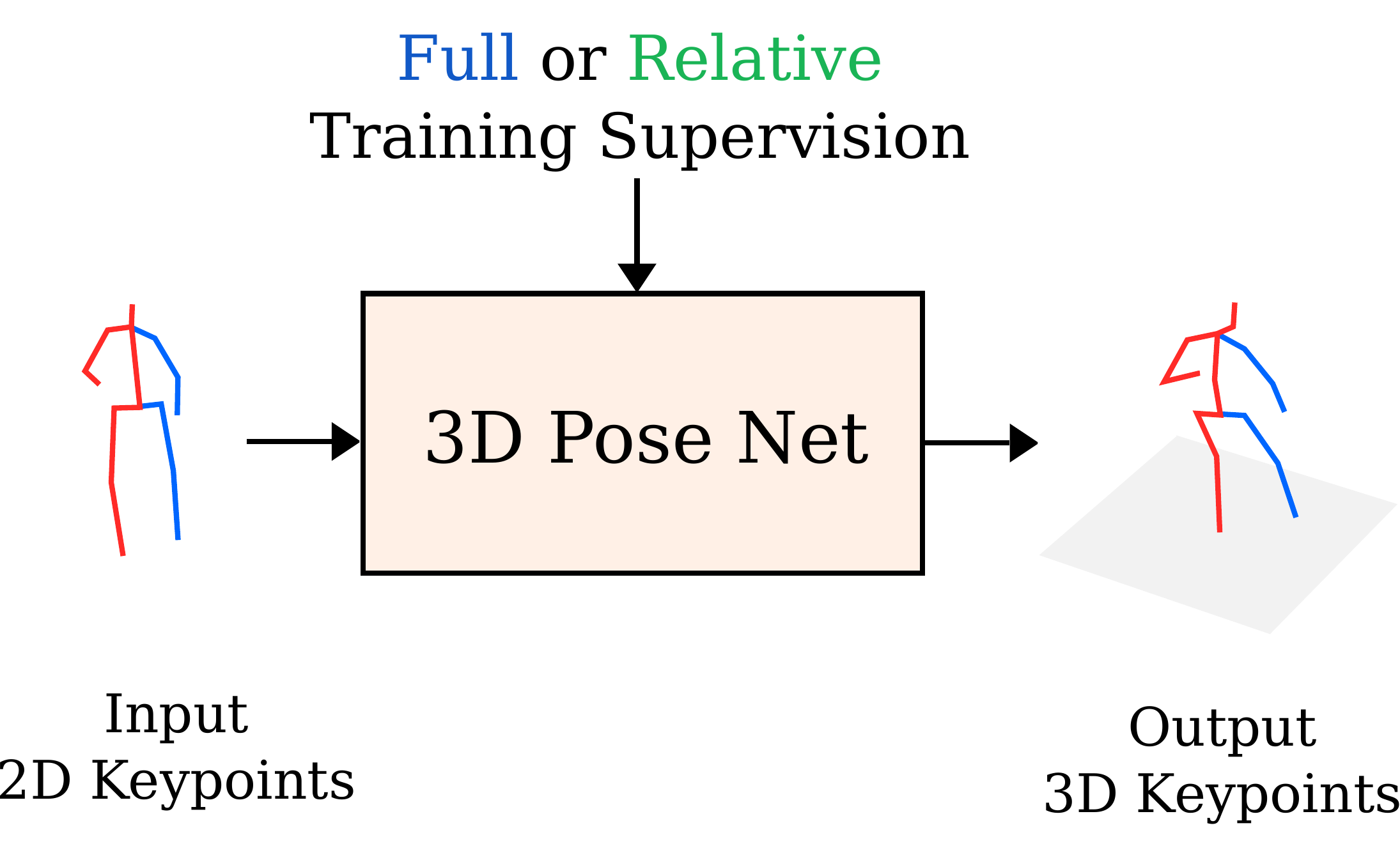}
    }~
    \subfigure[Full 3D Labels]{
        \centering
        \includegraphics[trim={0px 0px 0px 0px},clip,height=.18\textwidth]{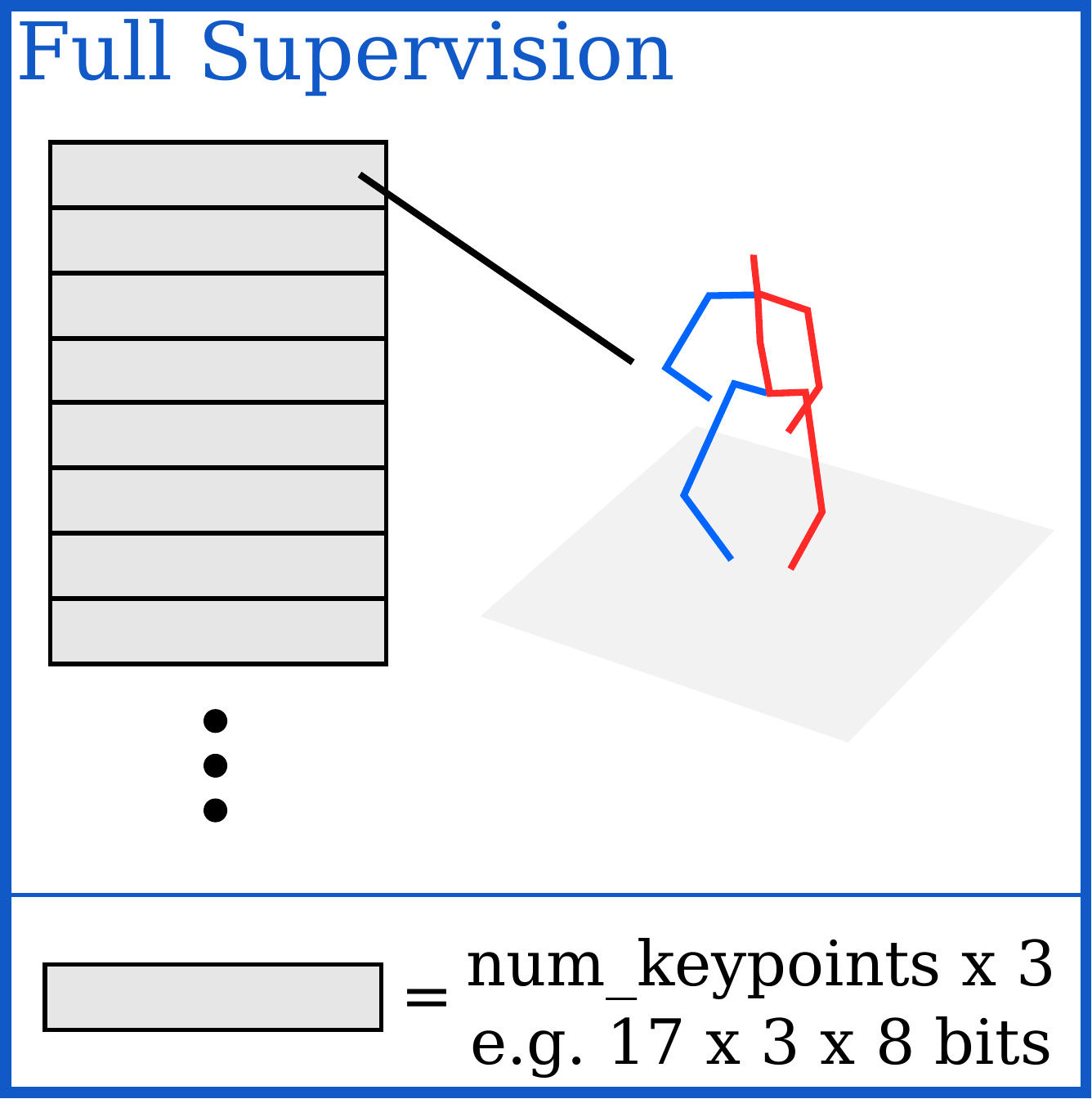}
    }~
     \subfigure[Relative Labels]{
       \centering
        \includegraphics[trim={0px 0px 0px 0px},clip,height=.18\textwidth]{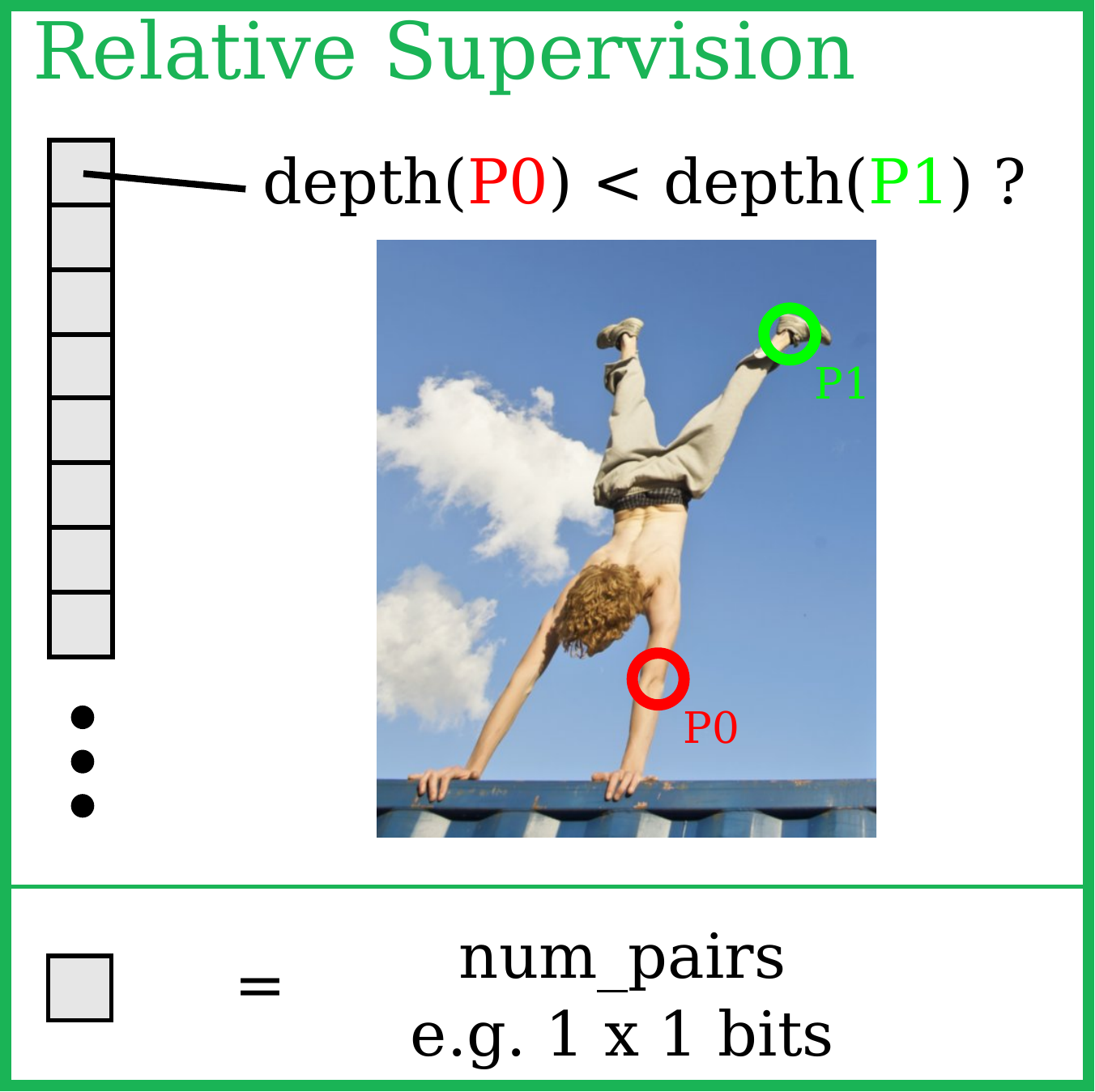}
    }~
    \subfigure[Train Data vs Error]{
        \centering
        \includegraphics[trim={0px 0px 0px 0px},clip,height=.18\textwidth]{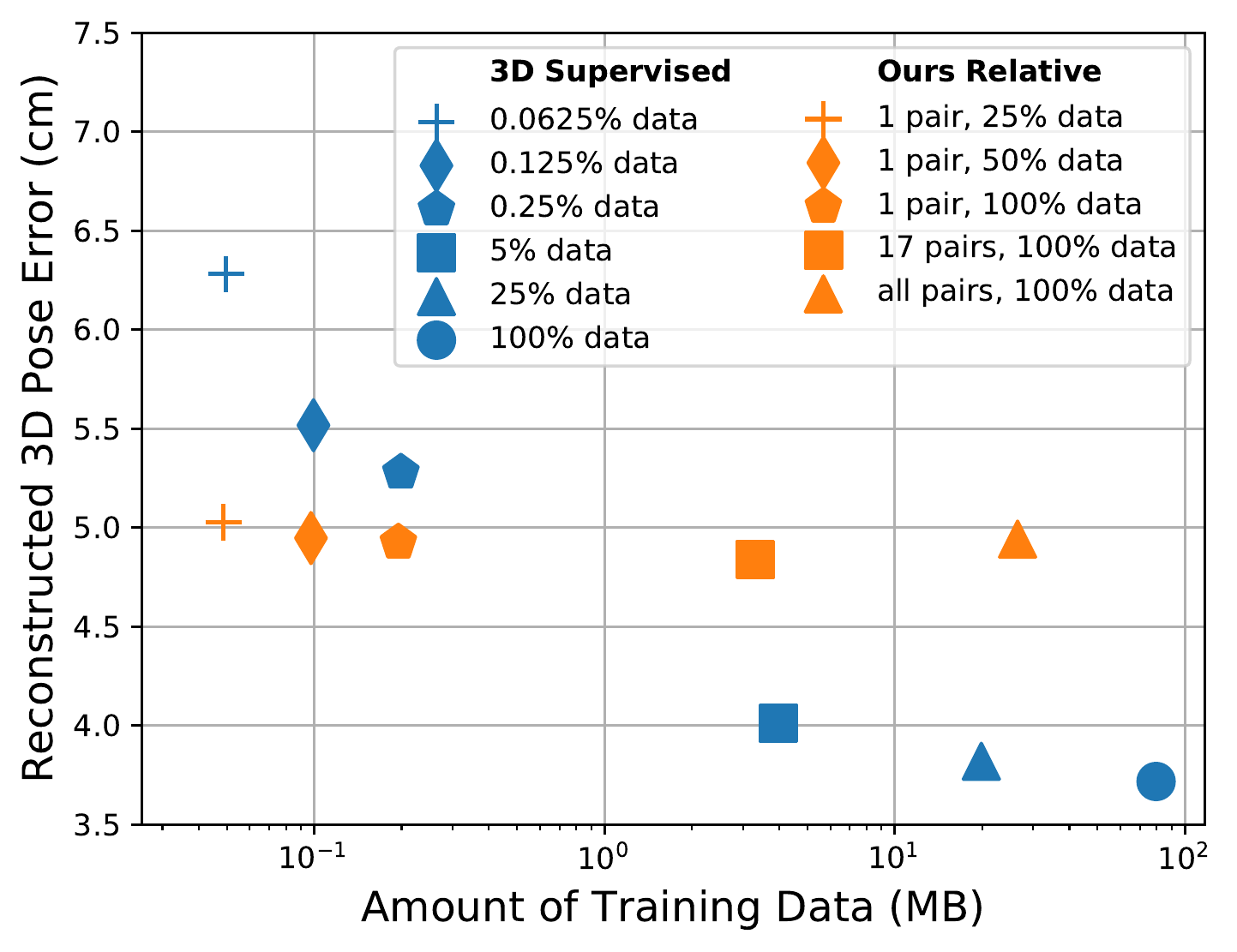}
    }
    \caption{(a) Lifting based methods take a set of 2D keypoints as input and predict their 3D position. (b) This is achieved using ground truth 3D poses during training. (c) We show that weak supervision specifying the relative depth of as little as one pair of keypoints per image is effective for training 3D pose estimation algorithms. (d) Our model predicts accurate 3D poses even with small amounts of relative training data, here measured in megabytes.}
    \label{fig:intro}
    \vspace{-5mm}
\end{figure*}

Reasoning about the pose of humans in images and videos is a fundamental problem in computer vision and robotics.
To ensure that future autonomous systems are safe to interact with, they need to be able to understand not only the positions but also the poses of the people around them.
Recent success in 2D pose estimation has been driven by larger, more varied, labeled datasets.
While laborious, it is possible for human annotators to click on the 2D locations of different body parts to generate such training data. 
Unfortunately, in the case of 3D pose estimation, it is much more challenging to acquire large amounts of training data containing people in real world settings with their corresponding 3D poses. 
This lack of large scale training data makes it difficult to both train deep models for 3D pose estimation and to evaluate the performance of existing methods in situations where there are large variations in scene types and poses. 
As a result, researchers have resorted to various alternative methods for collecting 3D pose training data - including motion capture, synthetic datasets, video, and multi-camera setups. 
In this work, we argue that instead of using additional hardware to acquire full 3D ground truth training data in controlled settings, Fig.~\ref{fig:intro} (b), we can make use of human annotated relative depth information from images in the wild, Fig.~\ref{fig:intro} (c).

Our main contributions are:
(1) a loss for 3D pose estimation of articulated objects that can be trained on sparse and easy to collect relative depth annotations with performance comparable to the state of the art,
(2) an empirical evaluation of the ability of crowd annotators to provide relative depth supervision in the context of human poses, 
and (3) a dataset of relative joint depth annotations that can be used for both training and evaluation purposes.

%%%%%%%%% RELATED WORK
\vspace{-2mm}
\section{Related Work} \label{sec:related_work}

{\bf 2D Pose Estimation:} Current state of the art methods for 2D human keypoint estimation are based on deep networks \cite{tompson2014joint,wei2016convolutional,newell2016stacked,cao2017realtime,he2017mask} trained on large quantities of supervised data \cite{Johnson10,sapp2013modec,andriluka20142d,lin2014microsoft}.
In addition to more sophisticated network architectures, a large driver in the improved accuracy of these approaches is the increase in the size and complexity \cite{ronchi2017benchmarking} of datasets that contain images with corresponding keypoint annotations indicating the locations, in pixels, of specific body parts.
%These annotations typically consist of a set of keypoints indicating the locations, in pixels, of specific body parts. Traditionally, these annotations cover a sparse set of locations such as joints but recently richer 2D annotations, enabling denser prediction, have been explored~\cite{guler2018densepose}.
While suitable for 2D pose estimation, by and large, most existing 2D datasets do not contain any supervision signal for 3D pose estimation.  
In this work we show that existing 2D pose datasets can indeed be used for 3D pose estimation by augmenting them with relative depth annotations collected from crowd annotators.\\

\noindent{}{\bf 3D Pose Estimation:} There exist two main categories of methods for 3D pose estimation: (1) end-to-end models and (2) lifting based approaches. 
The first set of models take a 2D image or the output of a person detector as input, and then produce an estimate of the individual's 3D pose. 
This is achieved by learning to regress the 3D keypoint locations during training, either as a set of 3D coordinates \cite{li20143d} or as volumetric heat maps \cite{pavlakos2017coarse}.
These methods assume the availability of a training set of 2D images paired with corresponding 3D annotations.
To further constrain the problem, it is possible to enforce a strong prior on the predictions in the form of a parameterized model of human body shape \cite{anguelov2005scape,LoperSMPL2015,tekin2016structured,tome2017lifting}. 
While this ensures realistic looking outputs \cite{kanazawa2017end}, it can be limiting if the prior is not flexible enough to cover the full range of valid poses.

As an alternative, lifting based approaches simply take a set of predicted 2D keypoints as input and \emph{lift} them into 3D.
The rise in popularity of these methods is driven by two factors: (1) the 2D location of keypoints is a strong cue indicating their 3D configuration and (2) the limited number of `in the wild' datasets featuring \emph{paired} 2D images with 3D poses. %forces the creative use of \emph{unpaired} 2D and 3D training data.
A variety of lifting approaches have been proposed that either frame the problem as one of regression \cite{moreno20173d,MartinezAEstimation}, data driven retrieval \cite{yasin2016dual,chen20173d,rogez2018lcr}, dictionary based reconstruction \cite{akhter2015pose}, or use generative adversarial learning \cite{yang20183d}.

Instead of requiring full 3D pose information for each individual in an input image, we propose a method that only needs a small amount of sparse data indicating the relative depth of different body parts. 
This results in high quality predictions with as little as one relative depth constraint per pose at training time.\\

\noindent{}{\bf 3D Pose Training Data:} A major factor holding back progress in 3D pose estimation is the lack of \emph{in the wild} datasets featuring images with ground truth 3D poses.
Most existing 3D pose datasets feature single individuals captured in controlled studio settings \cite{sigal2010humaneva,ionescu2014human3} and are challenging to acquire due to the need for specialized equipment such as motion capture cameras and markers. % body suits with markers.  
Setups with multiple cameras make it easier to capture small numbers of interacting people \cite{Joo_2015_ICCV,mehta2017monocular}, but require multiple synchronized cameras in confined spaces to produce accurate depth.
Depth cameras can be used to generate 3D training data \cite{shu2016learning}, but are usually limited to the indoors.
Alternative approaches use additional equipment such as inertial sensors \cite{pons2011outdoor,trumble2017total} or passive markers \cite{wang2011practical}.
The main limitation of these setups is that it is very difficult to also capture 2D images that cover all the variation in appearance that one would encounter in real world, non-studio, settings.

One technique to extend studio captured motion capture data is to use computer graphics techniques to generate potentially unlimited amounts of synthetic training data \cite{chen2016synthesizing,varol2017learning}.
Synthetic training data has been successful for low-level vision tasks such as depth and optical flow estimation \cite{mayer2016large}.
However, rendering realistic environments featuring people interacting with each other and their surroundings is a challenging problem. 
%For 3D pose, the goal is to generate images that look realistic both in terms of the rendered human's appearance and how they are interacting with their surroundings. 
Furthermore, even if it is possible to successfully generate plausible scenes, these methods are still limited by the variation in pose and the range of subject interactions that are present in the input motion capture data. 
%Augmenting the 3D body shapes to increase shape variation is possible, but pose retargeting is still a challenging problem.
Different graphical user interfaces have been explored to allow crowd annotators to annotate 3D pose information in existing image datasets. Examples include, manually configuring 3D skeletons \cite{nguyen2010annotation} or providing coarse body part orientation information \cite{maji2011large,maji2011action,andriluka20142d}.
These can be very laborious tasks, taking a large amount of time per image.\\

\noindent{}{\bf Relative Depth Supervision:}
Using ordinal relations between a sparse set of point pairs as a weak form of supervision has been previously explored in the context of dense depth estimation in monocular images \cite{zoran2015learning,Chen2016Single-ImageWild}. While this is not comparable to metric ground truth depth, it enables the easy collection of data that can be used for both training and evaluating monocular depth estimation.
This type of data has also been used for 3D pose estimation \cite{taylor2000reconstruction,bourdev2009poselets}. However, unlike previous work that require complete annotations of all joint pairs to infer the 3D pose, we use only sparse relative annotations. Our annotations, Fig.~\ref{fig:intro} (c), specify the relative distance between keypoints and the camera. Also \cite{pons2014posebits} crowdsources sparse pairwise pose constraints. However, in their case the relative relationships are in the coordinate frame of the person in the image, and used as a replacement for 2D keypoints. We show that depth annotations relative to the camera are easy to collect and can be combined with existing 2D keypoint annotations for improving 3D pose estimation.

In parallel to our work, \cite{pavlakos2018ordinal} also explored training 3D pose estimation models using relative (i.e. ordinal) depth constraints. They use multiple relative annotations per image to fine-tune 3D supervised trained models. In contrast, we use as little as one relative pair per image and focus on the setting where no camera intrinsics or 3D supervised ground truth are available for pre-training or for calibrating the output of the model. Finally, we also conduct a detailed user study evaluating how accurate crowd annotators are at providing relative depth annotations for human poses.

%%%%%%%%% METHOD
\section{Method} \label{sec:models}
Our goal is to predict the 3D pose of an individual depicted in an input image. 
We represent pose in 2D as a set of coordinates $\mathbf{p} \in \mathbb{R}^{2 \times J}$, where each element $\mathbf{p}_{j} = [u_j, v_j]$ is a row vector that encodes the location, in pixels, of one of  $J$ different joints. 
For each $\mathbf{p}$, we aim to infer its position in 3D $\mathbf{P} \in \mathbb{R}^{3 \times J}$, where each entry specifies the location of the joint $j$ in 3D, $\mathbf{P}_{j} = [x_j, y_j, z_j]$.
In this work we take inspiration from lifting based approaches, Fig.~\ref{fig:intro} (a), and attempt to learn the parameters of a function $f: \mathbb{R}^{2\times J} \to \mathbb{R}^{3 \times J}$, that maps 2D input keypoints to their estimated 3D position, where $f(\mathbf{p}) = \mathbf{\hat{P}}$.
We parametrize $f$ as a neural network, where the input joint positions $\mathbf{p}$ can come from the output of a 2D human pose estimation algorithm \eg \cite{wei2016convolutional, newell2016stacked,cao2017realtime}.  

\subsection{Supervised 3D Pose Estimation}
Given a set of $N$ input 2D keypoints and their corresponding ground truth 3D pose one could use a supervised loss to train $f$
\begin{equation}
\mathcal{L}_{sup}(\mathbf{\hat{P}}, \mathbf{P}) = ||\mathbf{\hat{P}} - \mathbf{P}||_2.
\end{equation}
This is the approach taken in \cite{MartinezAEstimation}, where a neural network, $f$, is trained to project the input coordinates $\mathbf{p}$ into 3D.
While they only need to infer the missing $z_j$ values for each 2D keypoint, their model predicts each $[x_j, y_j, z_j]$ coordinate, making the approach more robust to errors in the input 2D locations. 

\subsection{3D Pose Estimation with Relative Constraints}
As noted earlier, acquiring large quantities of varied ground truth 3D pose data is challenging. 
Instead, we opt to use much weaker supervision in the form of depth ordering labels that describe the relative distance to the camera for a pair of keypoints, see Fig.~\ref{fig:intro} (c).

We assume we have access to a set of crowdsourced relative annotations for an image $i$, $\mathcal{A}^i = \{(j_1, k_1, r_1), (j_2, k_2, r_2), ..., (j_A, k_A, r_A)\}$, where each annotation $(j,k,r)$ is a tuple specifying the joints $j$ and $k$ and their estimated relative depth $r\in{\{-1,1,0\}}$.
The number of specified pairwise constraints, $A$, can be different for every image, and varies between one and $\binom{J}{2}$, when the ordinal supervision is provided for every pair of keypoints.
The value $r = -1$ indicates that $z_j < z_k + \epsilon$ (joint $j$ is closer to the camera compared to $k$), while $r = 1$ specifies that $z_k < z_j + \epsilon$. If the distance between two keypoints is below a certain tolerance $\epsilon$, then $r = 0$. In practice, this corresponds to the case in which human annotators cannot disambiguate the relative position of two keypoints.
Unless otherwise noted, we explore the setting where $r\in{\{-1,1\}}$.

%
% ranking and root
%
Similar to \cite{Chen2016Single-ImageWild}, we use a pairwise ranking loss to encourage our model to predict the correct depth ordering of a 3D keypoint pair
\begin{equation}\label{eqn:rel}
\mathcal{L}_{rel}(\mathbf{\hat{P}}, \mathcal{A}) = \sum_{(j,k,r) \in \mathcal{A}}
\begin{cases}
    \log(1 + \exp(-r\hat{d}_{jk})), & r = -1, +1\\
    ||\hat{d}_{jk}||_2, & r = 0,
\end{cases}
\end{equation}
where $\hat{d}_{jk} = \lambda{}(\hat{z}_j - \hat{z}_k)$, $\hat{z}_j$ being the predicted depth from our network for keypoint $j$, and $\lambda{}$ controlling the strength of the loss.
In practice, we found that it is important to normalize the range of depth values $\hat{d}$ to ensure numerical stability~\cite{wang2017learning}.
This is achieved by scaling by the mean absolute depth difference across each minibatch during training.
We also constrain our 3D predictions so they are centered at a root joint that is encouraged to remain at the origin 
\begin{equation}
\mathcal{L}_{root}(\mathbf{\hat{P}}) = ||\mathbf{\hat{P}}_{root}||_2.
\end{equation}
This controls the range of the output space, as the network does not have to model all possible poses at all possible distances from the camera.

%
% reprojection
%
The above ranking loss only encourages the relative distances to the camera to be respected for each keypoint pair, in essence constraining the $z$ values.
To force the correct location in both $x$ and $y$ in image space, we use a reprojection loss
\begin{equation}\label{eqn:scaled_ortho}
\mathcal{L}_{proj}(\mathbf{\hat{P}}, \mathbf{p}, \mathbf{v}, s) = \sum_{j}||v_j(\Pi\mathbf{\hat{P}}_j^{\top} - \mathbf{p}_j)||_2,
\end{equation}
where $v_j \in{\{0, 1\}}$ is a visibility flag and $\Pi$ is a projection matrix.
If no camera intrinsic information is available $\Pi = \left[\begin{smallmatrix}s & 0 & 0\\ 0 & s & 0\end{smallmatrix}\right]$, \ie scaled orthographic projection, and in addition to predicting the 3D pose, our network also learns to predict the scaling parameter $s$ for each input pose.
If the ground truth focal lengths are available during training we can use perspective projection, and the reprojection loss becomes 
\begin{equation}\label{eqn:weak_persp}
\mathcal{L}_{proj}(\mathbf{\tilde{P}}, \mathbf{p}, \mathbf{v}, s) = \sum_{j}||v_j(\Pi\mathbf{\tilde{P}}_j^{\top} - \mathbf{p}_j)||_2.
\end{equation}
where $\Pi = \left[\begin{smallmatrix} f_x & 0 & 0\\ 0 & f_y & 0\end{smallmatrix}\right]$, and $\mathbf{\tilde{P}}_j = \left[x_j / (z_j + s), y_j / (z_j + s), 1 \right]$. Now the network's scaling parameter $s$ has a different interpretation and is used to predict the distance from the camera to the center of the person in 3D. 

% If the ground truth focal lengths are available during training $\Pi = \left[\begin{smallmatrix} f_x & 0 & 0\\ 0 & f_y & 0\end{smallmatrix}\right]$ and $\mathbf{\tilde{P}}_j = \left[x_j / (z_j + s), y_j / (z_j + s), 1 \right]$.
% Now the network's scaling parameter, $s$, has a different interpretation, where it is used to predict the distance from the camera to the center of the person in 3D, and the reprojection loss becomes 
% \begin{equation}\label{eqn:weak_persp}
% \mathcal{L}_{proj}(\mathbf{\tilde{P}}, \mathbf{p}, \mathbf{v}, s) = \sum_{j}||v_j(\Pi\mathbf{\tilde{P}}_j^{\top} - \mathbf{p}_j)||_2.
% \end{equation}

%
% skeleton
%
Even with the above terms, 3D pose estimation from 2D inputs is heavily underconstrained as many different 3D pose configurations can respect both the relative depth constrains and the reprojection loss. 
To further constrain the problem we include one additional geometric loss that enforces weak prior knowledge related to the ratio between the lengths of the different limbs. 
We assume we are given an input skeleton $\mathcal{B} = \{(b_1^1,b_1^2, l_1), (b_2^1,b_2^2, l_2), ... , $ $(b_B^1,b_B^2, l_B)\}$, consisting of $B$ `bones' (\ie limbs), where each entry $(b^1,b^2, l)$ specifies the indices of the keypoint pair that are the endpoints for that particular limb, and its length $l$. 
The limb length loss then measures the difference in length between the predicted limb and the predefined reference length,
\begin{equation}\label{eqn:skel}
\mathcal{L}_{skel}(\mathbf{\hat{P}}) = \sum_{(b^1,b^2,l) \in \mathcal{B}}
||\text{len}(\mathbf{\hat{P}}, b^1,b^2) - l||_2,
\end{equation}
where $\text{len}(\mathbf{\hat{P}}, j, k) = ||\mathbf{\hat{P}}_j - \mathbf{\hat{P}}_k||_2$. 
In practice we do not minimize the difference between the absolute bone lengths but instead normalize the predicted and reference bones by fixing one of the limbs to be unit length, in effect constraining their ratios as in \cite{wang2014robust,zhou2017towards}.
The skeleton loss also implicitly enforces symmetry between the sets of left and right limbs. % \eg left and right forearms.

Our final loss $\mathcal{L}$ is the combination of the above four terms with additional weighting hyperparameters to control the influence of each component
\begin{equation}
\mathcal{L} = \frac{1}{N}\sum_{i=1}^N\mathcal{L}_{root}(\mathbf{\hat{P}}^i) + \alpha\mathcal{L}_{rel}(\mathbf{\hat{P}}^i, \mathcal{A}^i) + \beta\mathcal{L}_{proj}(\mathbf{\hat{P}}^i, \mathbf{p}^i, \mathbf{v}^i, s^i) + \gamma\mathcal{L}_{skel}(\mathbf{\hat{P}}^i).
\end{equation}

%%%%%%%%% EXPERIMENTS
\section{Human Relative Depth Annotation Performance}
Our model for 3D pose estimation makes use of relative depth annotations at training time. 
In order to use this type of supervision, it is important to understand how accurately can humans provide these labels. 
This is in contrast to the study carried out in \cite{marinoiu2016pictorial}, which investigates the ability of humans to observe and physically reenact a target 3D pose. 
To measure the quality of relative joint annotations collected via a crowd sourcing platform we performed an evaluation using participants recruited on Mechanical Turk with 1,000 images randomly selected from the Human3.6M dataset~\cite{ionescu2014human3}, as it features ground truth depth. 
An image of the user interface for the task is provided in the supplementary material.
For each annotation task, the crowd annotators were presented with an image along with two randomly selected keypoints and were instructed to imagine themselves looking through the camera and report which of the two keypoints appeared closer to them. 
We forced annotators to choose from one of the two possibilities and did not provide a `same distance' option for ambiguous situations, as those cases can be inferred by inspecting the disagreement between annotators.
For each of the 1,000 images, we collected five random pairs of keypoints, ensuring that five different annotators labeled the same keypoints and image combination. 
In total, this resulted in 25,000 individual annotations collected from 301 annotators, with an average of 83 annotations each. 
We merged each of the five votes per keypoint pair using the crowd annotation system of \cite{branson2017lean}, resulting in a single predicted label per pair.
We found this to perform slightly better than a majority vote, with the added benefit of providing a probabilistic label. 

\begin{figure*}[t]
    \centering
    \subfigure[Accuracy vs Depth]{  
        \centering
        \includegraphics[trim={10px 10px 10px 31px},clip,width=.21\textwidth]{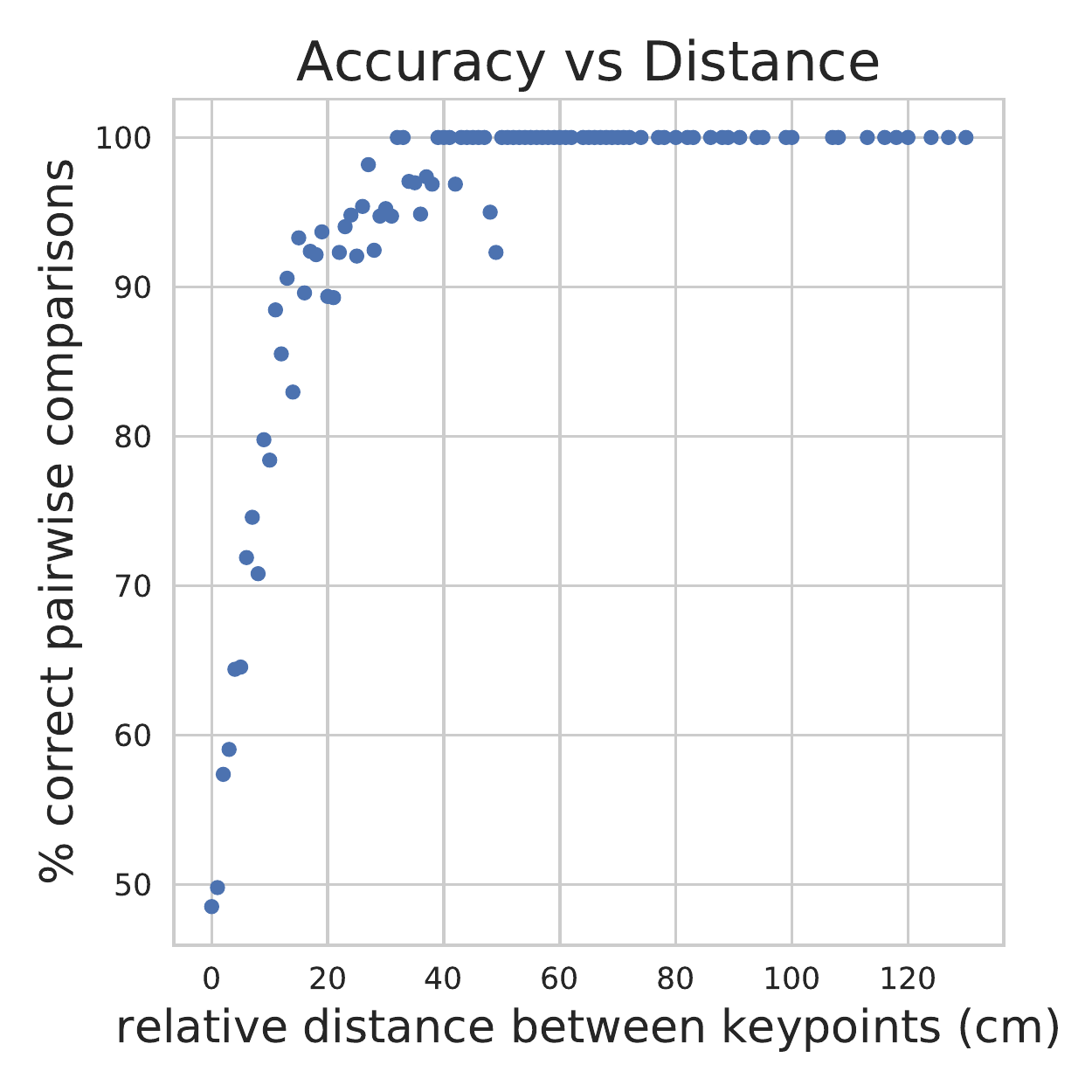}
    }~
     \subfigure[Annotator Accuracy]{
       \centering
        \includegraphics[trim={10px 10px 10px 31px},clip,width=.21\textwidth]{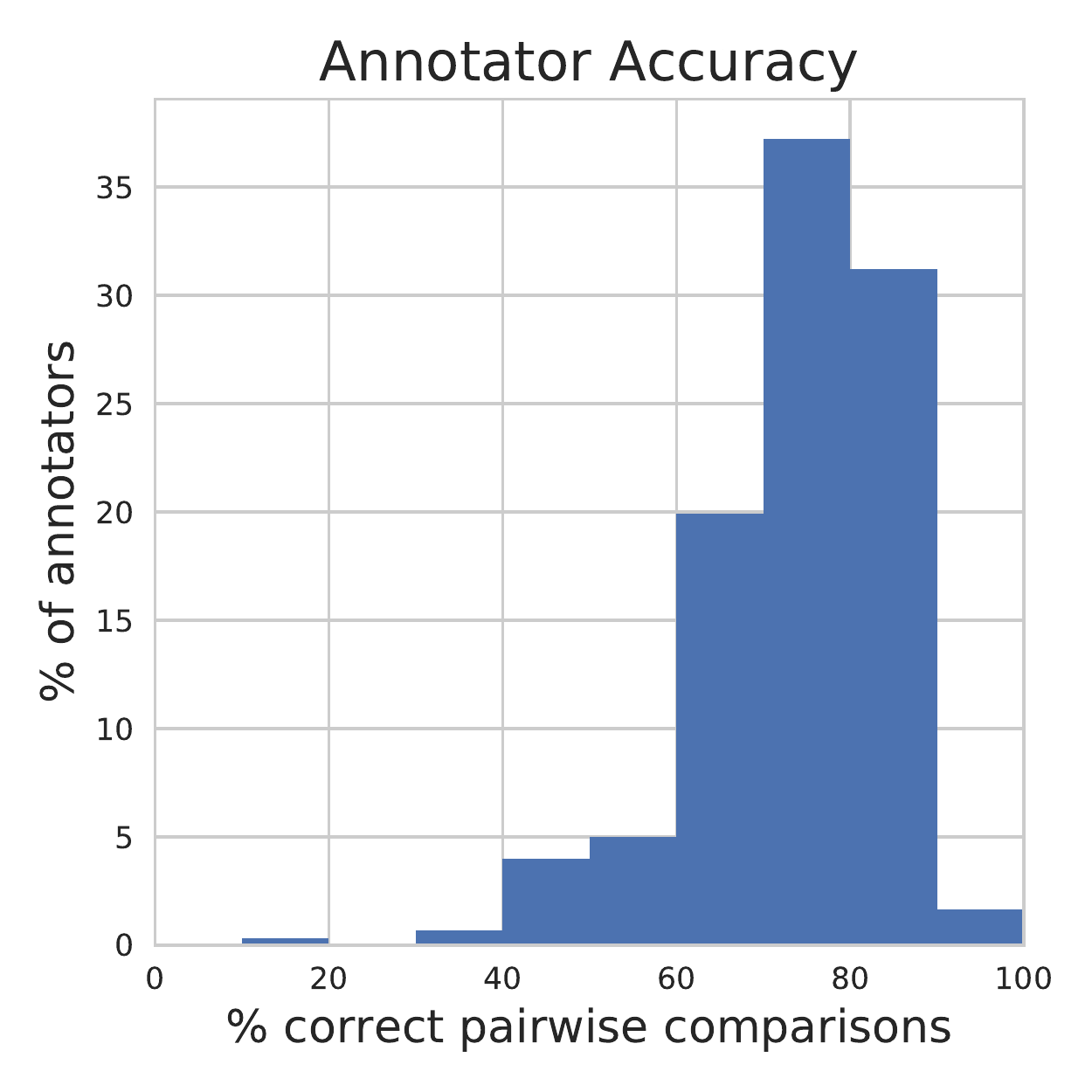}
    }~
    \subfigure[Annotator Skill]{
        \centering
        \includegraphics[trim={10px 10px 10px 30px},clip,width=.21\textwidth]{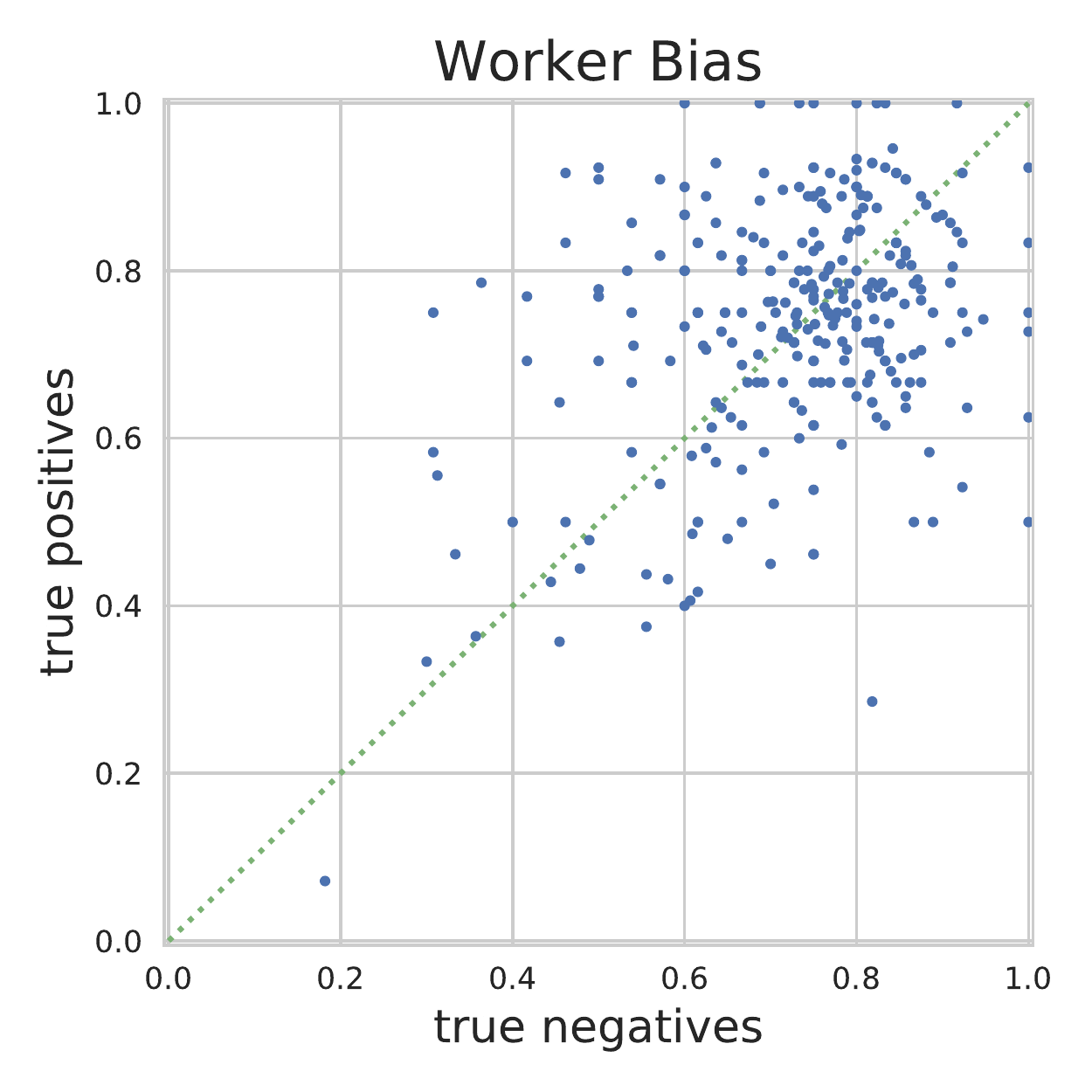}
    }~
    \subfigure[Image Difficulty]{
        \centering
        \includegraphics[trim={10px 10px 10px 31px},clip,width=.21\textwidth]{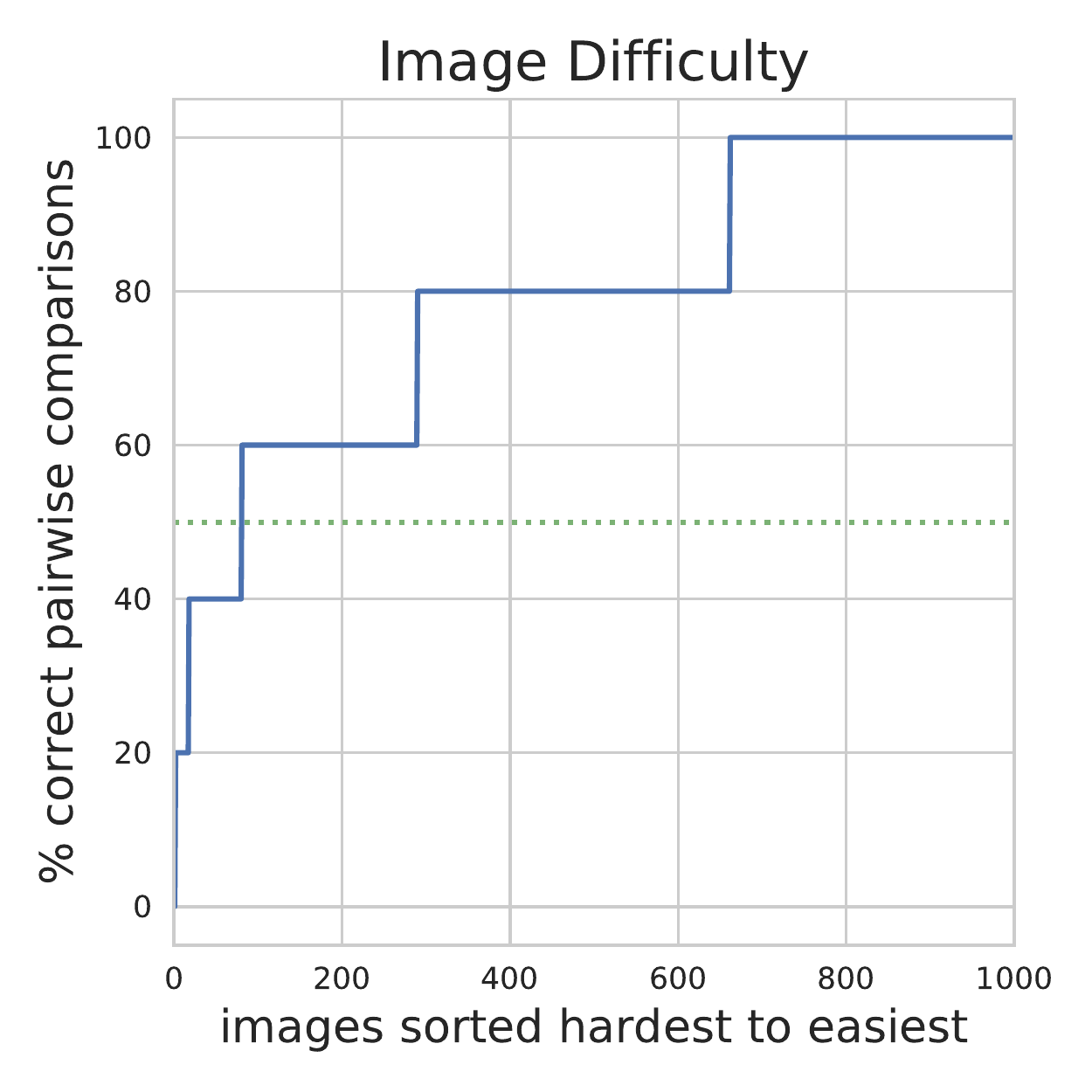}
    }
    \caption{(a-c) Analysis of human performance for the relative keypoints depth annotation task measured on 1,000 images from Human3.6M. Annotators always perform better than chance and are more than 90\% accurate when the distance between pairs is larger than 20 cm. (d) Sorting of the images based on their overall keypoint difficulty.}
    \label{fig:user_study}
\end{figure*}

\begin{figure}
\centering
\includegraphics[trim={135px 15px 110px 0px},clip,width=1.0\textwidth]{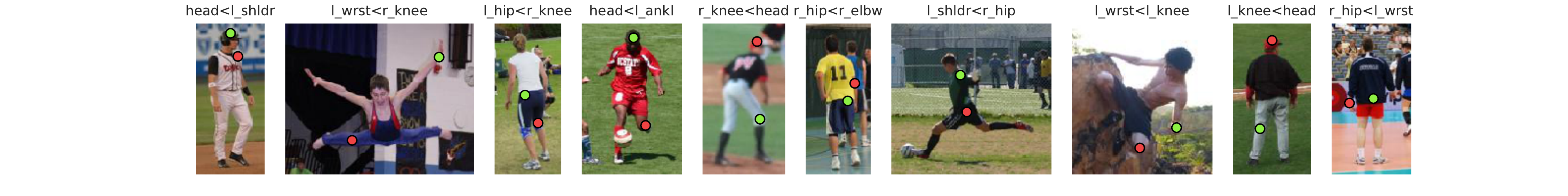}
\caption{Examples of the supervision collected for the LSP dataset. The names of the keypoints are written on top and the green keypoint is annotated to be closer to the camera compared to the red one. Images are sorted left to right from most to least confident based on the annotator's agreement, and the last two examples illustrate particularly challenging cases where the keypoints are at a similar distance to the camera.}
\label{fig:lsp_turk}
\end{figure}

\newpage
We observed a bias in the annotations due to the fact that crowd annotators tend to not factor in the forward lean of the camera in Human3.6M when making their predictions. 
Correcting for this bias during evaluation results in 71\% of the total raw annotations being correct, and 79\% of the relative orderings being correct after merging the five individual annotations using \cite{branson2017lean}. 
Note that fixing this bias requires ground truth depth data, and is performed here to evaluate annotators' performance. 
In real world scenarios our model would learn this annotation bias. 
A more detailed exploration of the bias, and its correction, is presented in the supplementary material.

In Fig.~\ref{fig:user_study} we quantify the accuracy of the crowd annotations, shown after correcting for the annotation bias. 
In Fig.~\ref{fig:user_study} (a) we see that for keypoint pairs that are separated by more than 20 cm our merged predictions are correct over 90\% of the time, where random guessing is 50\%. While only a small number of annotators annotated over 90\% of the pairs correctly, Fig.~\ref{fig:user_study} (b), the vast majority tend to perform better than random guessing. In Fig.~\ref{fig:user_study} (c) we observe that the rate of true positives versus true negatives for every annotator is fairly symmetric, indicating that annotators are equally good at providing the correct answer independently of a keypoint being in front or behind another one. Some image and keypoint combinations are more challenging than others, and in Fig.~\ref{fig:user_study} (d) we sort the images from hardest to easiest based on the percentage of keypoint pairs that are correctly annotated. For over two thirds of the images, four out of the five pairs are correctly annotated. 
Importantly, the cases where annotators have trouble predicting the correct keypoint order, by and large, tend to be ambiguous pairs where the actual ground truth distances are small. 
These results indicate that human annotators can indeed provide high quality weak 3D pose information in the form of relative depth annotations that can be used as supervision for our algorithm.

Using the same protocol described above we collected annotations for all 2,000 images in the Leeds Sports Pose (LSP) dataset \cite{Johnson10}. LSP features a much larger variation in camera viewpoint and pose compared to Human3.6M. 
Again, we selected five random keypoint pairs per image, with five repeats, resulting in a total of 50,000 annotations. 
Annotations were performed by 348 annotators who provided an average of 144 labels each.
Example annotations after merging the five responses using \cite{branson2017lean} can be seen in Fig.~\ref{fig:lsp_turk}.
Unlike Human3.6M, there is no ground truth depth data available for LSP, so to evaluate the quality of the crowd annotations two of the authors independently annotated the same subset of $500$ keypoint pairs.
Agreement between the two was 84\%, where the majority of disagreements occurred in ambiguous cases.
For the set of pairs where the two annotators agreed, the merged crowd annotations were the same 90.2\% of the time. 
These results are consistent with the performance on Human3.6M, despite the larger variation in poses and viewpoints.

%%%%%%%%%%%%%%%%%%%%%%%%%%%%%%%%%%%%%%%%
% TABLE WITH ABLATION STUDY ON HUMAN3.6M
%
\begin{table}
\centering
\scalebox{0.65}{
\begin{tabular}{l|llll|cc}
\hline
\textbf{Model}         & \textbf{}            & \textbf{}         & \textbf{}             & \textbf{}                   & \multicolumn{2}{c}{\textbf{3D Pose Error (mm)}} \\ \hline
\textbf{}              & \textbf{Reprojection Type} & \textbf{Skeleton} & \textbf{Amount Train} & \textbf{Distance Tolerance} & \textbf{Scale} & \textbf{Procrustes} \\ \cline{1-7}
     1) 3D Supervised          &  -                   & -                 & -                     & -                    
            & 49.62 & 36.54 \\ \hline
     2) \textbf{Ours Relative} & Scaled Orthographic  & H36 Avg           & 1 Pair                & No                          & 64.58 & 48.97 \\ \hline
     3) Known focal length      & Perspective          & H36 Avg           & 1 Pair                & No                          & 65.22 & 48.54 \\ %\cline{2-7}
     4) Generic skeleton     & Scaled Orthographic  & Avg from \cite{taylor2000reconstruction}       & 1 Pair                & No                          & 73.26 & 55.27 \\
            %            & Scaled Orthographic  & H36 Subject       & 1 Pair                & No                          & 65.35 & 49.10 \\ %\cline{2-7}
     5) No skeleton loss        & Scaled Orthographic  & H36 Avg           & 1 Pair                                           & No                 & 118.34 & 89.81 \\
     6) Amount train      & Scaled Orthographic  & H36 Avg           & All Pairs (136)            & No                         & 63.96 & 48.90 \\ 
     7) Distance tolerance       & Scaled Orthographic  & H36 Avg           & 1 Pair                & Yes (100mm)                 & 63.45 & 47.54 \\  \hline
\multicolumn{7}{c}{ } \\
\end{tabular}
}
\caption{Ablation study of our model on Human3.6M (H36). The last two columns show the test time performance when an optimal single parameter re-scaling (\textit{Scale}) or full rigid alignment (\textit{Procrustes}) is performed for each predicted pose based on the ground truth. `3D Supervised' is our re-implementation of \cite{MartinezAEstimation}. `Ours Relative' shows the default settings adopted for our model, while rows 3-7 contain different variants.}
% \caption{Results for different variants of our model on Human3.6M (H36). The last two columns show the test time performance when an optimal single parameter re-scaling (\textit{Scale}) or full rigid alignment (\textit{Procrustes}) is performed for each predicted pose based on the ground truth. The row \textit{Ours Relative} shows the default settings adopted for our model. The row \textit{Ours Relative} shows the default settings adopted for our model.}
\label{tab:ablation_h36}
\vspace{-3mm}
\end{table}

%%%%%%%%% EXPERIMENTS
\section{3D Pose Estimation Results}
We use a similar model to \cite{MartinezAEstimation} for all our experiments and report the 3D pose estimation performance of our model on the Human3.6M~\cite{ionescu2014human3} and LSP~\cite{Johnson10} datasets in the following sections. A description of our network architecture and implementation details are available in the supplementary material, along with additional qualitative results.

\vspace{-3mm}
\subsection{Human3.6M}
\label{sec:resh36}
As noted in \cite{chen20173d}, many different evaluation protocols have been defined for Human3.6M, making it very challenging to comprehensively compare to all existing methods. 
We opt to use protocol \#2 from the state-of-the-art \cite{MartinezAEstimation}, as it is the model most similar to ours, and has been used by several recent approaches.
Here, training is performed on subjects [1, 5, 6, 7, 8] and the test set consists of all frames and cameras for subjects [9, 11]. 
Some baselines, evaluate at 10fps which we observe to make little difference to the test scores.
As in \cite{MartinezAEstimation}, Procrustes alignment is performed at test time between each prediction and its corresponding test example.
With the exception of the results in Table~\ref{tab:protocol_2}, where we average across actions, everywhere else we report results by averaging across all frames. 
Finally, unless specified, we use 17 2D keypoints as input and predict their corresponding 3D locations, with the relative annotations derived from the ground truth depth.

%%%%%%%%%%%%%%%%%%%%%%%%%%%%%%%%%%%%%%%%%%%%%%%%%
% FIGS SHOWING AFFECTS OF NOISE
%
\begin{figure*}[t]
    \centering
    \subfigure[Overall Error]{ 
        \centering
        \includegraphics[trim={0px 0px 0px 0px},clip,height=.175\textwidth]{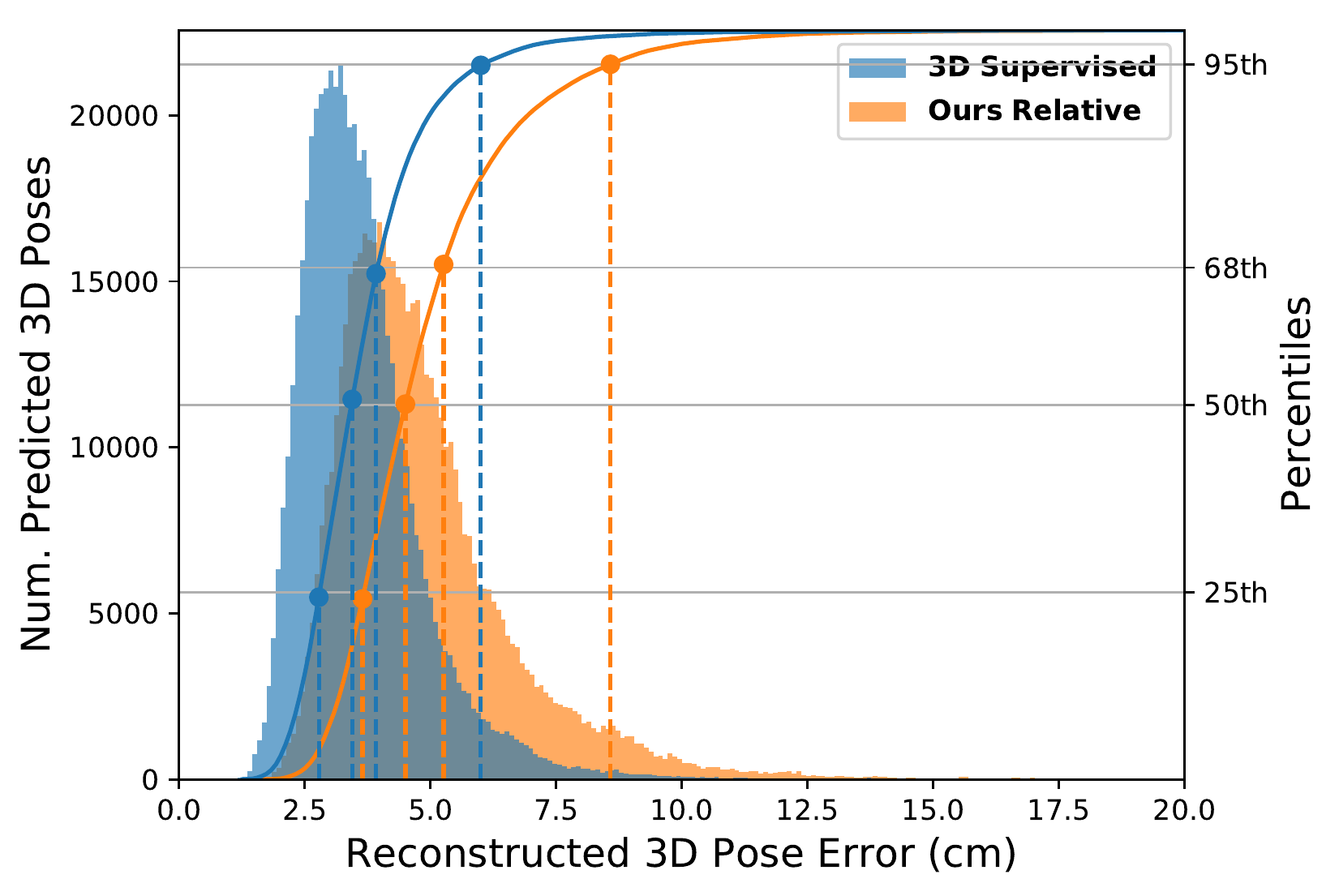}
    }~
    \subfigure[Per Keypoint Error]{
        \centering
        \includegraphics[trim={0px 0px 0px 0px},clip,height=.175\textwidth]{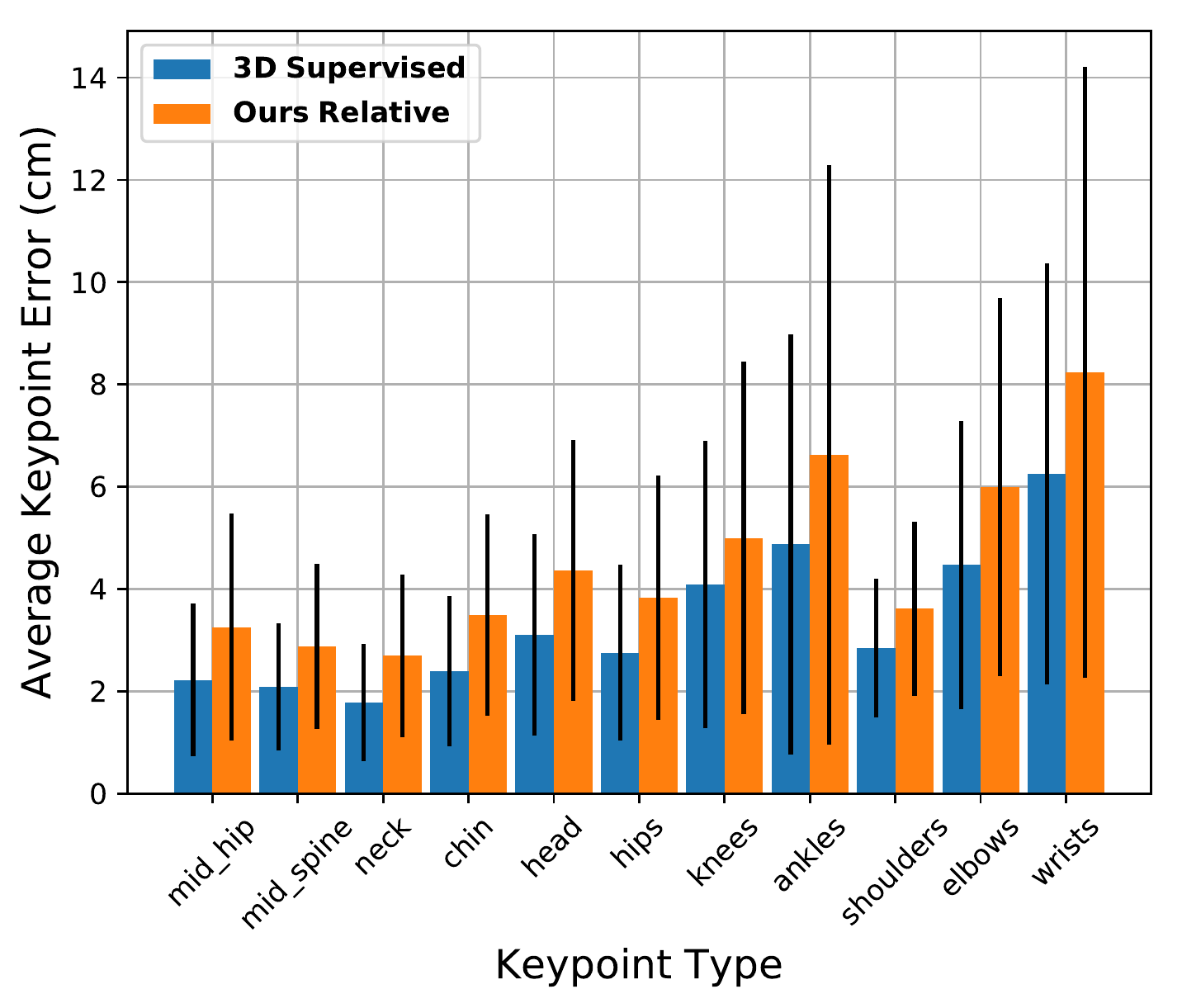}
    }~
     \subfigure[2D Keypoint Noise]{
       \centering
        \includegraphics[trim={0px 0px 0px 0px},clip,height=.175\textwidth]{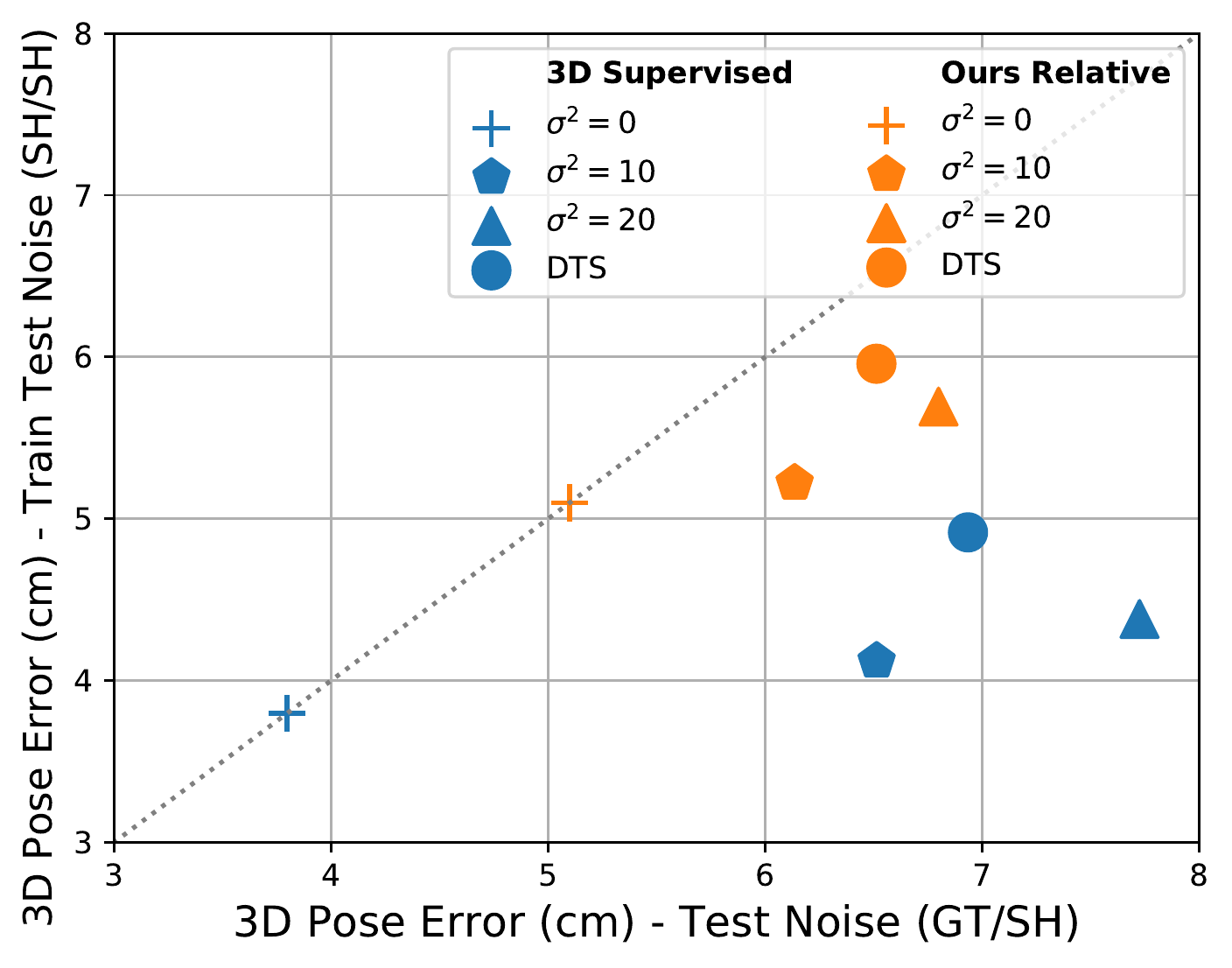}
    }~
    \subfigure[3D Annotation Noise]{
        \centering
        \includegraphics[trim={0px 0px 0px 0px},clip,height=.175\textwidth]{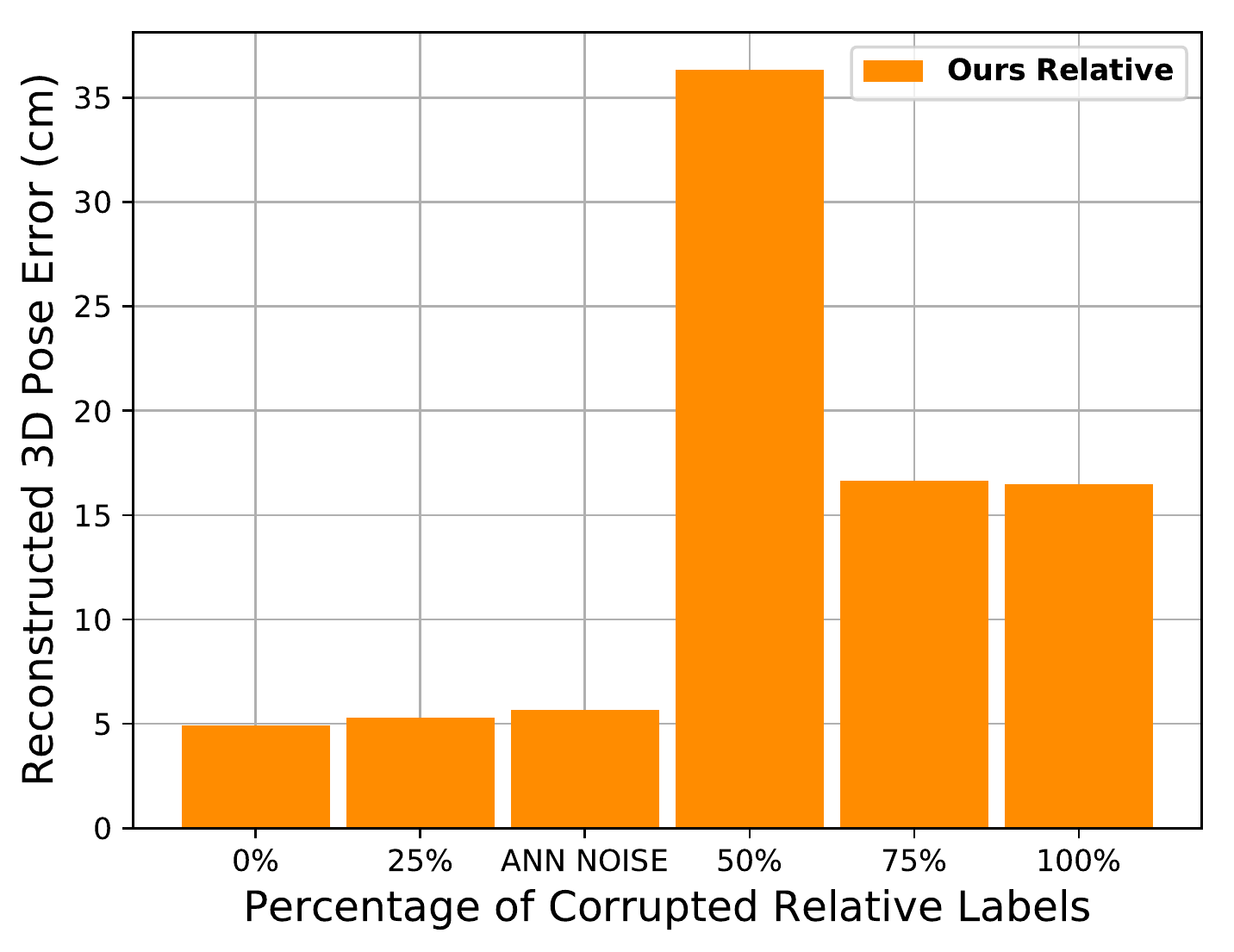}
    }
    \caption{(a-b) Histogram of the errors on the Human3.6M dataset (lines and dots respectively show the cumulative sum and percentiles) with a breakdown over individual keypoints. (c-d) Our method is robust to increasingly strong perturbations on the 2D input keypoint locations at test time and noise on the relative depth labels during training.}
    \label{fig:h36_abalation}
    \vspace{-4mm}
\end{figure*}

We explored different configurations of our model at training time, by varying the type of reprojection loss adopted, the proportions of the input skeleton, the amount of training data, and the distance tolerance. 
We describe below our most interesting findings from the results in Table~\ref{tab:ablation_h36}. 
\textit{Reprojection}: The availability of ground truth focal lengths at training time, Eqn.~\ref{eqn:weak_persp}, yields comparable performance to using the scaled orthographic projection of Eqn.~\ref{eqn:scaled_ortho}.
\textit{Skeleton}: Using less accurate limb length ratios in Eqn.~\ref{eqn:skel}, obtained from \cite{taylor2000reconstruction} instead of Human3.6M's training set average, slightly hurts the accuracy, while entirely removing the skeleton loss significantly reduces performance.
\textit{Amount Train}: Increasing the amount of relative depth pairs does not significantly alter the results. However, this effect is likely due to the high redundancy present in Human3.6M. This is consistent with the findings of Fig.~\ref{fig:intro} (d), where we report the error against the percentage of training images used.
\textit{Distance Tolerance}: Setting the depth tolerance to 100 mm helps performance. This is because using $r=0$ in Eqn.~\ref{eqn:rel} forces the network to constrain a pair of predicted keypoints to be at the same depth, as opposed to just determining their relative order. 

Some of these configurations are more realistic than others when using crowd provided annotations on an `in the wild' image dataset. We denote `Ours Relative' as the model with the most realistic assumptions and use it for the remaining experiments.

Fig.~\ref{fig:h36_abalation} summarizes the overall performance and robustness to noise of our model.
In Fig.~\ref{fig:h36_abalation} (a) we show a histogram of the pose errors on the test set both for our method and ~\cite{MartinezAEstimation}. The mode and median of the two curves are 10mm from each other. 
However, our method suffers from more catastrophic errors, as can be seen in the longer tail. 
This is due to the fact that even when respecting all the relative depth labels we do not fully constrain their absolute depth. 
This can also be seen in the breakdown of the error over keypoint type in Fig.~\ref{fig:h36_abalation} (b). 
As one might expect, body extremities such as ankles and wrists show a larger error (and deviation). 
Fig.~\ref{fig:h36_abalation} (c) shows the degradation in performance for the cases in which the 2D input keypoints are obtained by adding a Gaussian noise $\mathcal{N}(0,\,\sigma^{2})$ with increasingly high variance (up to 20) to the 2D ground truth keypoints or by using the outputs of a keypoint detector~\cite{newell2016stacked} fined-tuned on the Human3.6M training set, taken from \cite{MartinezAEstimation}. The performance of~\cite{MartinezAEstimation} is better when the \emph{train} and \emph{test} data have the same amount of noise degradation (lower error along the y axis), while our method performs best when noise is only added at \emph{test} time (lower error along the x axis). We hypothesize this behavior is due to the presence of the reprojection and skeleton losses at training time, which encourages our method to find plausible 3D poses that respect the input poses, making it more robust to slight changes in the distribution of the input keypoints.
In Fig.~\ref{fig:h36_abalation} (d) we demonstrate that our model is also robust to noise in the relative depth labels during \emph{training}. Performance is mostly unchanged when up to 25\% of the labels are randomly flipped. The third bar corresponds to the amount of noise obtained from simulated crowd annotators, regressed from Fig.~\ref{fig:user_study} (a). This is of interest, as it shows performance with noise comparable to what we would expect to collect in the wild. The worst performance is obtained when the labels are randomly flipped, and improves for cases in which the amount of noise is larger than 50\%, as the model is able to exploit structure that is still present in the data, but produces poses that are flipped back to front.
Finally, in Table~\ref{tab:protocol_2} we compare our model to existing fully 3D supervised approaches. 
Even with significantly less training data, and without any architecture exploration, we still perform competitively. When available, 3D ground truth is a very powerful training signal, but our results show that relative depth data can still be used at the expense of little accuracy at test time. 
Our model is robust to using noisy predicted 2D keypoints at test time, again with a minor decrease in performance.  
Example 3D predicted poses on Human3.6M can be seen in the supplementary material.

%%%%%%%%%%%%%%%%%%%%%%%%%%%%%%%%%%%%%%%
% TABLE COMPARING TO OTHERS ON HUMAN3.6M
% Table adapted from \cite{hossain2017exploiting}
% Have removed lines that are not directly applicable
\begin{table*}[t]
\footnotesize
\tabcolsep=0.6mm
%\vspace{-3mm}
\centering
\scalebox{0.75}{
\begin{tabular}{@{}lcccccccccccccccc@{}}
\hline
 & Direct. & Discuss & Eating & Greet & Phone & Photo & Pose & Purch. & Sitting & SitingD & Smoke & Wait & WalkD & Walk & WalkT & Avg\\
\hline
%Akhter \& Black~\cite{akhter2015pose}* 14j & 199.2 & 177.6 & 161.8 & 197.8 & 176.2 & 186.5 & 195.4 & 167.3 & 160.7 & 173.7 & 177.8 & 181.9 & 176.2 & 198.6 & 192.7 & 181.1\\
%Ramakrishna et al.~\cite{ramakrishna2012reconstructing}* 14j & 137.4 & 149.3 & 141.6 & 154.3 & 157.7 & 158.9 & 141.8 & 158.1 & 168.6 & 175.6 & 160.4 & 161.7 & 150.0 & 174.8 & 150.2 & 157.3\\
%Zhou et al.~\cite{zhou2016sparseness}*  14j & 99.7 & 95.8 & 87.9 & 116.8 & 108.3 & 107.3 & 93.5 & 95.3 & 109.1 & 137.5 & 106.0 & 102.2 & 106.5 & 110.4 & 115.2 & 106.7\\
%Bogo et al.~\cite{bogo2016keep}*  14j & 62.0 & 60.2 & 67.8 & 76.5 & 92.1 & 77.0 & 73.0 & 75.3 & 100.3 & 137.3 & 83.4 & 77.3 & 86.8 & 79.7 & 87.7 & 82.3\\ \hline
%Rogez et al.~\cite{rogez2016mocap} 13j 1/64 fps  & -- & -- & -- &-- & -- & -- & --  & -- & -- & -- & -- & -- & -- & -- &-- & 87.3\\
%Nie et al.~\cite{nie2017monocular} 13j    & 62.8 & 69.2 & 79.6 & 78.8 & 80.8 & 86.9 & 72.5 & 73.9 & 96.1 & 106.9 & 88.0 & 70.7 & 76.5 & 71.9 & 73.2 & 79.5\\
%Mehta et al.~\cite{mehta2016monocular}  14j  & -- & -- & -- & -- & -- & -- & -- & -- & -- & -- & -- & -- & -- & -- & -- & 54.6\\
%Moreno-Noguer~\cite{moreno20173d}  14j & 67.5 & 79.0 & 76.5  &  83.1  &  97.4  &  74.6  &  72.0  &  102.4   &  116.7  & 87.7 & 100.8  &  94.6  &  75.2  &  82.7  &  74.9  &  85.6\\ % separate per action
Sanzari et al.~\cite{sanzari2016bayesian} & 48.8 & 56.3 & 96.0 & 84.8 & 96.5 & 105.6 & 66.3 & 107.4 & 116.9 & 129.6 & 97.8 & 65.9 & 130.5 & 92.9 & 102.2 & 93.2\\
Rogez et al.~\cite{rogez:hal-01505085} & -- & -- & -- & -- & -- & -- & -- & -- & -- & -- & -- & -- & -- & -- & -- & 71.6\\
%Kanazawa et al.~\cite{kanazawa2017end} unpaired & -- & -- & -- & -- & -- & -- & -- & -- & -- & -- & -- & -- & -- & -- & -- & 67.5\\
Kanazawa et al.~\cite{kanazawa2017end}  & -- & -- & -- & -- & -- & -- & -- & -- & -- & -- & -- & -- & -- & -- & -- & 58.1\\
Pavlakos et al.~\cite{pavlakos2017coarse}  & 47.5 & 50.5 & 48.3 & 49.3 & 50.7 & 55.2 & 46.1 & 48.0 & 61.1 & 78.1 & 51.1 & 48.3 & 52.9 & 41.5 & 46.4 & 51.9\\
Tekin et al.~\cite{tekin2017learning}   & -- & -- & -- & -- & -- & -- & -- & -- & -- & -- & -- & -- & -- & -- & -- & 50.1\\
Fang et al.~\cite{fang2017learning}   & 38.2 & 41.7 & 43.7 & 44.9  & 48.5  & 55.3 & 40.2 & 38.2 & 54.5 & 64.4 & 47.2 & 44.3 & 47.3 & 36.7 & 41.7 & 45.7\\
Hossain et al.~\cite{hossain2017exploiting} (T) & 36.9&  37.9&	42.8&	40.3&	46.8&	46.7&	37.7&	36.5&	48.9&	52.6&	45.6&	39.6&	43.5&	35.2&	38.5&	42.0\\
Pavlakos et al.~\cite{pavlakos2018ordinal} (E) & 34.7 & 39.8 & 41.8 & 38.6 & 42.5 & 47.5 & 38.0 & 36.6 & 50.7 & 56.8 & 42.6 & 39.6 & 43.9 & 32.1 & 36.5 & 41.8\\
\hline 
Martinez et al.~\cite{MartinezAEstimation}  17j GT/GT & -- & -- &	-- &	-- &	-- & -- &	-- &	-- &	-- &	-- &	-- &	-- &	-- &	-- &	-- &	37.1\\
Martinez et al.~\cite{MartinezAEstimation}  17j SH/SH & 39.5 & 43.2 &	46.4&	47.0&	51.0&56.0&	41.4&	40.6&	56.5&	69.4&	49.2&	45.0&	49.5&	38.0&	43.1&	47.7\\
\hline
3D Supervised 17j GT/GT & 31.1 & 37.0 & 34.3 & 36.3 & 37.2 & 42.5 & 36.6 & 33.8 & 39.9 & 49.3 & 37.0 & 37.7 & 38.8 & 33.1 & 37.5 & 37.5\\
{\bf Ours Relative} 17j GT/GT & 43.6 & 45.3 & 45.8 & 50.9 & 46.6 & 55.3 & 43.3 & 47.3 & 56.6 & 74.3 & 47.1 & 48.5 & 52.1 & 48.5 & 49.8 & 50.3\\
{\bf Ours Relative} 17j GT/GT (N) & 46.4 & 47.3 & 49.7 & 53.3 & 52.3 & 59.2 & 43.8 & 50.9 & 80.5 & 107.9 & 54.3 & 51.6 & 53.1 & 52.4 & 53.3 & 57.1\\
\hline
3D Supervised 16j SH/SH & 40.2 & 44.4 & 46.6 & 48.3 & 53.3 & 56.7 & 41.7 & 42.4 & 55.9 & 68.1 & 50.8 & 46.2 & 51.8 & 40.4 & 45.5 & 48.8\\ 
{\bf Ours Relative} 16j SH/SH & 51.1 & 50.8 & 56.4 & 60.5 & 60.6 & 64.7 & 48.9 & 52.1 & 67.5 & 89.0 & 59.1 & 55.8 & 61.3 & 59.2 & 59.0 & 59.7\\ 
{\bf Ours Relative} 16j SH/SH (N) & 54.0 & 54.5 & 70.4 & 64.0 & 78.3 & 71.6 & 50.4 & 58.8 & 113.9 & 131.0 & 76.7 & 60.1 & 64.8 & 61.8 & 63.7 & 71.6\\ \hline
\multicolumn{17}{c}{ } \\
\end{tabular}
}
\caption{3D pose reconstruction errors in mm on Human3.6M using protocol \#2 as defined in Sec.~\ref{sec:resh36}. GT and SH are the ground truth and fine-tuned (from \cite{MartinezAEstimation}) detected 2D input keypoints respectively, with `j' indicating the number of joints used for testing. (T) represents the use of temporal information, (E) extra training data, and (N) the use of simulated annotation noise in the relative depth pairs.}
\label{tab:protocol_2}
\vspace{-5mm}
\end{table*}

% protocol 2 - all frames of [9,11] for test with rigid alignment

% Bogo et al~\cite{bogo2016keep}              NO only cam 3
% Akhter \& Black~\cite{akhter2015pose}       NO only cam 3 ??
% Zhou et al~\cite{zhou2016sparseness}        NO only cam 3 ?? 
% Ramakrishna et al~\cite{ramakrishna2012reconstructing}  NO only cam 3 ??

% Rogez et al~\cite{rogez2016mocap}           YES except 1/64
% Nie et al~\cite{nie2017monocular}           NO only sub 11
% Mehta et al~\cite{mehta2016monocular}       NOT CLEAR

% Moreno-Noguer~\cite{moreno20173d}           NO - their proto 1 trains sep on each action
% Pavlakos et al~\cite{pavlakos2017coarse}    YES 
% Tekin et al~\cite{tekin2017learning}        YES
% Fang et al.~\cite{fang2017learning}         YES
% Hossain et al.~\cite{hossain2017exploiting} YES
% Martinez et al.~\cite{MartinezAEstimation}  YES

%%%%%%%%%%%%%%%%%%%%%%%%%%%%%%%%%%%%%
% LSP
%

\subsection{Leeds Sports Pose Dataset}
\begin{figure*}[t]
    \centering
    \subfigure{  
        \centering
        \includegraphics[trim={140px 50px 85px 55px},clip,width=.45\textwidth]{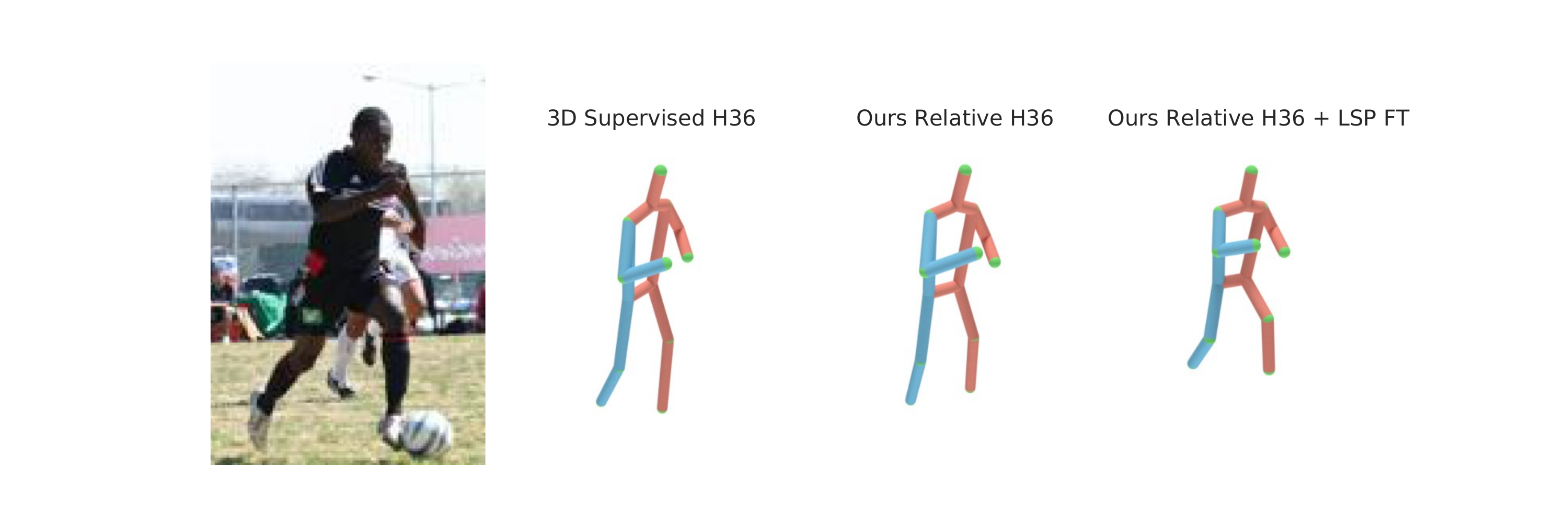}
    }~
    \subfigure{
        \centering
        \includegraphics[trim={140px 50px 85px 55px},clip,width=.45\textwidth]{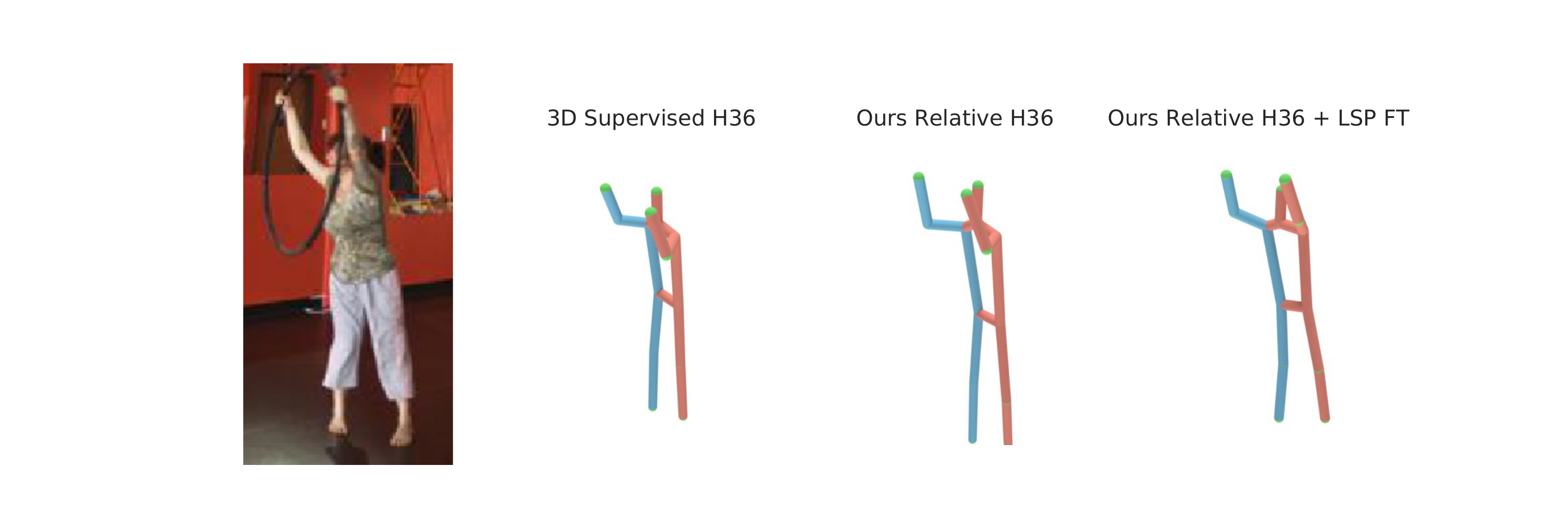}
    }\\
    \subfigure{  
        \centering
        \includegraphics[trim={140px 50px 85px 55px},clip,width=.45\textwidth]{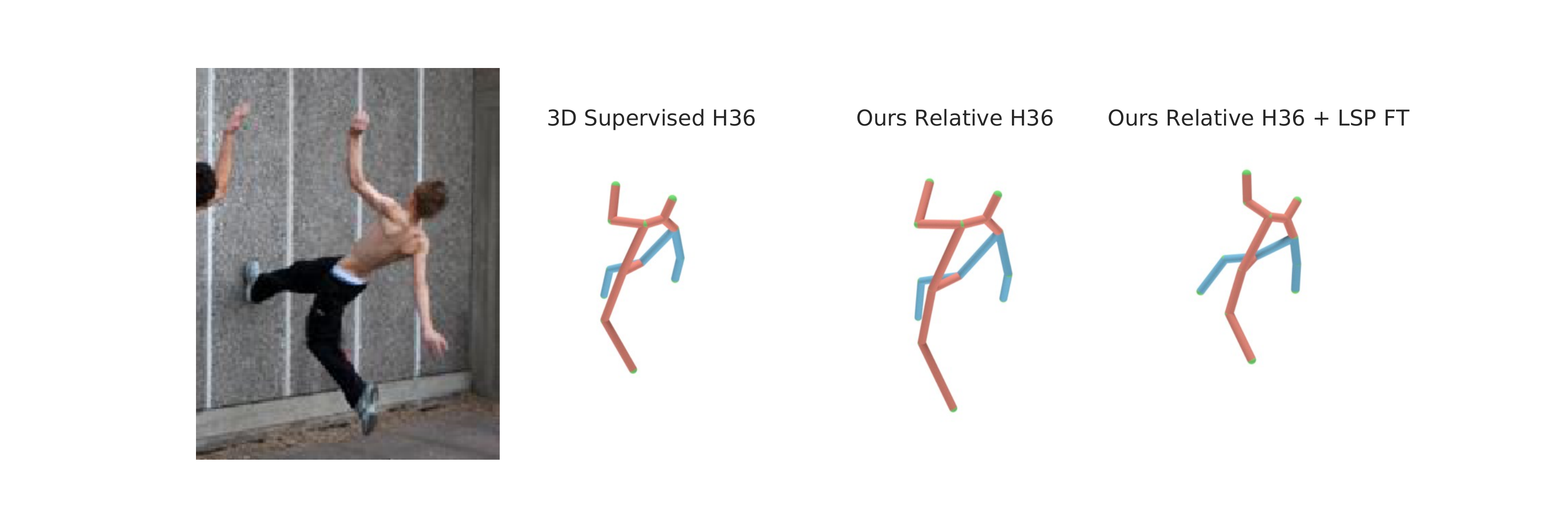}
    }~
    \subfigure{
        \centering
        \includegraphics[trim={140px 50px 85px 55px},clip,width=.45\textwidth]{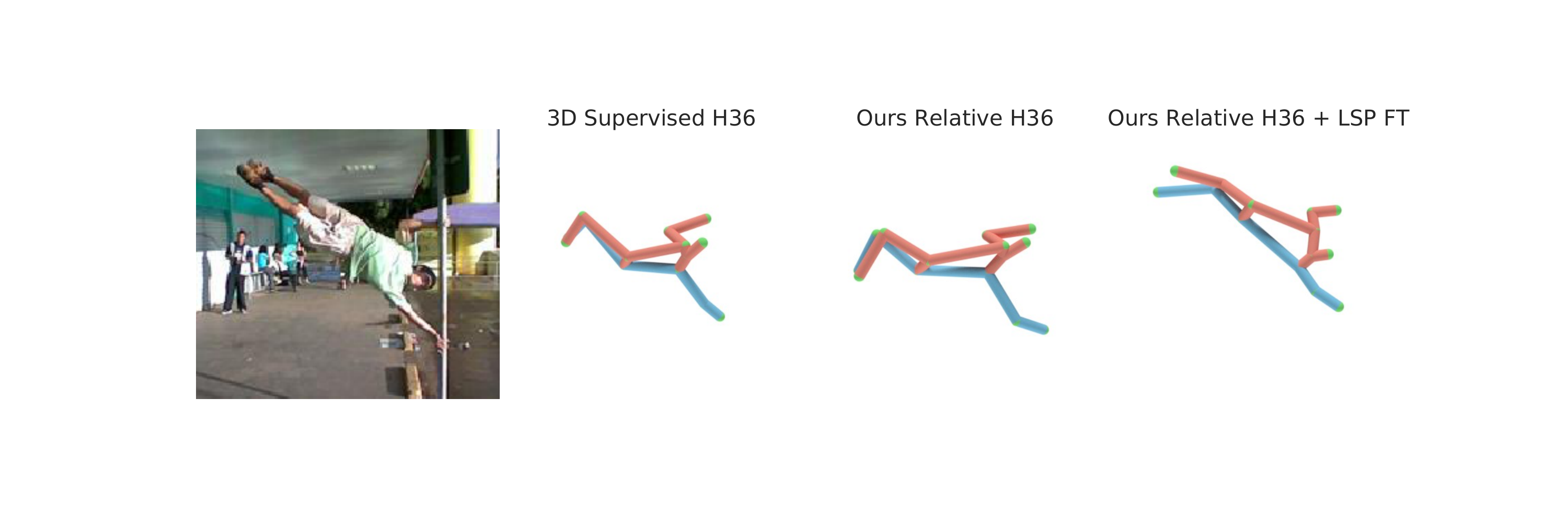}
    }\\
     \subfigure{
       \centering
        \includegraphics[trim={140px 50px 85px 55px},clip,width=.45\textwidth]{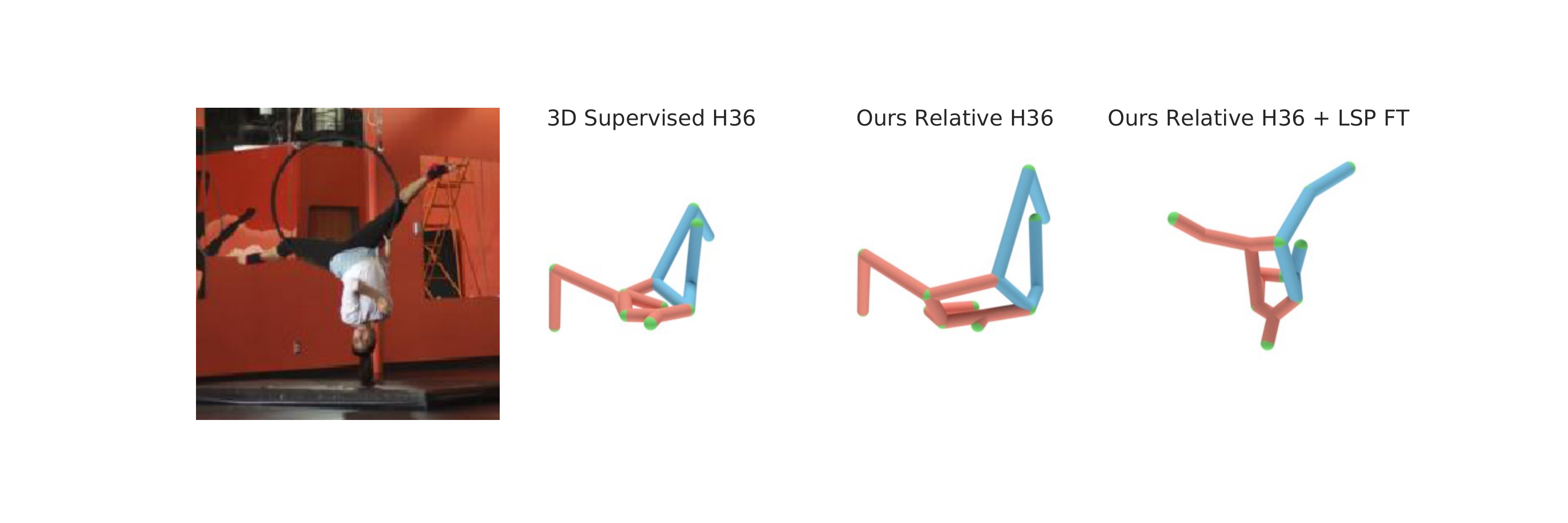}
    }~
    \subfigure{
        \centering
        \includegraphics[trim={140px 50px 85px 55px},clip,width=.45\textwidth]{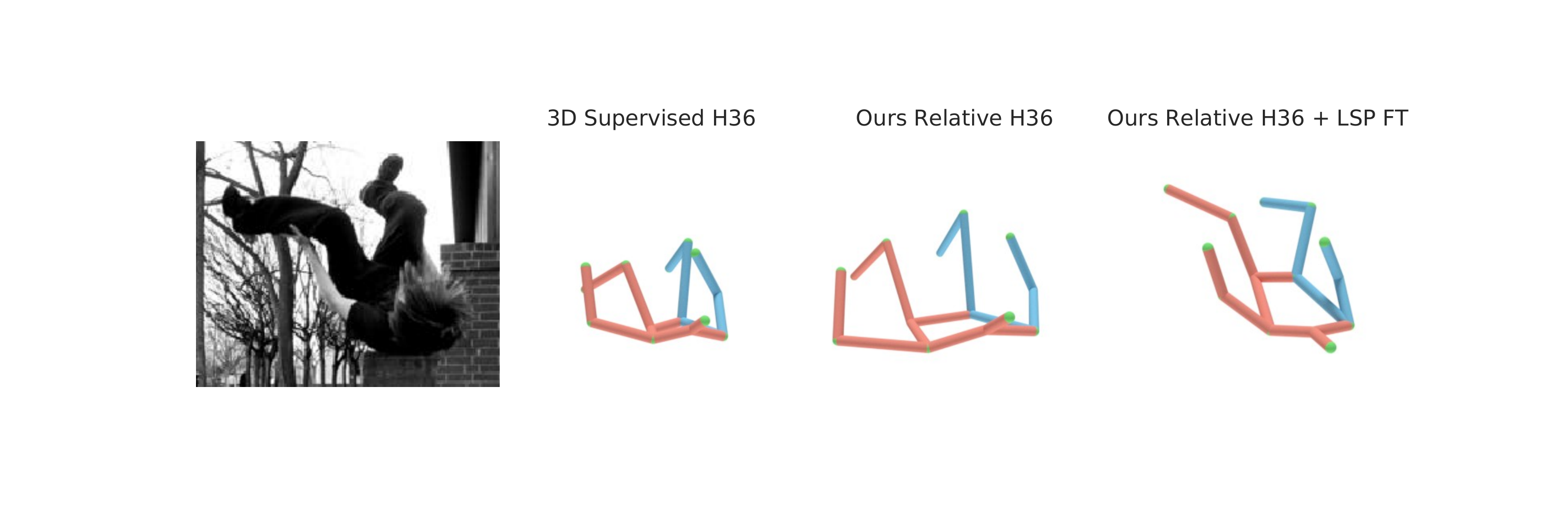}
    }
\vspace{-3mm}
\caption{Predicted 3D poses on LSP. Fine-tuning (FT) on LSP significantly improves the quality of our predictions especially for images containing uncommon poses and viewpoints that are not found in Human3.6M, such as those visualized in the last row.}
\label{fig:lsp_qual}
\vspace{-4mm}
\end{figure*}

We perform experiments on the LSP dataset which, unlike Human3.6M, does not contain ground truth 3D poses. 
As a result, we evaluate the models by measuring the percentage of relative labels incorrectly predicted compared to the merged crowd annotations collected via Mechanical Turk.
The 3D supervised model \cite{MartinezAEstimation} and our relative model, with one comparison per input pose, trained on Human3.6M achieve test errors of 34.4\% and 34.3\% respectively.
We are able to further reduce the error of our relative model to 24.1\% by fine-tuning it using the ordinal annotations collected for the LSP training set. 
Furthermore, training our relative model from scratch using exclusively the LSP training set also outperforms the supervised baseline obtaining an error of 27.1\%.
This shows that weak training data is sufficient for competitive performance on this benchmark, and we can successfully make use of noisy annotations to improve the prediction of 3D poses if 3D ground truth is available. 
Example outputs of the above methods on the LSP dataset are displayed in Fig.~\ref{fig:lsp_qual}.

% We perform experiments on the LSP dataset which, unlike Human3.6M, does not contain ground truth 3D poses. As a result, we train and test our model using the merged crowd annotations collected via Mechanical Turk.
% Pre-training the 3D supervised model \cite{MartinezAEstimation} and our relative model on Human3.6M, with one relative comparison per input pose, results in test errors of 34.3\% and 35.2\% respectively. 
% Here, test error refers to the percentage of relative labels incorrectly predicted compared to the merged crowd annotations. 
% Our model uses only weak training data but is still competitive compared to the fully supervised baseline.
% However, if we fine-tune our model on our crowd collected LSP training set, the error reduces to 25.6\%.
% Example predicted 3D poses are displayed in Fig.~\ref{fig:lsp_qual}.
% This shows that we can successfully make use of noisy training annotations, improving the output 3D poses.

%%%%%%%%% CONCLUSION
\vspace{-4mm}
\section{Conclusion}

%%%%%%%%%%%%%%%%%%%%%%%%%%%%%%%%%%%%%%%%%%%%%%%%%
% LSP QUALITATIVE
% <left> <lower> <right> <upper>

% wrap-up
We presented a weakly supervised approach for 3D human pose estimation. 
We showed that sparse constraints that indicate the relative depth of pairs of keypoints can be used as a training signal, resulting in competitive results at a fraction of the amount of training data.
Unlike most approaches that require ground truth 3D poses, our method can be applied to legacy image collections, as only the input 2D keypoints and relative depth annotations are required. 
This opens the door to using existing datasets for 3D pose estimation in the wild.

% future work
Large scale annotation is time consuming and expensive, even when only collecting weak supervision. 
In future, we plan to investigate efficient, active learning based, approaches for collecting annotations \cite{liu2017active}. 
Current state-of-the-art 2D pose estimation algorithms perform best on single humans in isolation and their performance deteriorates when there are large numbers of occluded keypoints and closely interacting people \cite{ronchi2017benchmarking}.
Including weak 3D information for multiple interacting individuals may help resolve some of these ambiguities.\\

%%%%%
\small{\noindent\textbf{Acknowledgements:} Google for their gift to the Visipedia project and AWS for research credits.}

\bibliography{egbib}

%%%%%%%%% SUPPLEMENTARY
\newpage
\section{Supplementary Material}

\subsection{Implementation Details}
\begin{figure*}[ht]
    \centering
    \includegraphics[trim={0px 0px 0px 0px},clip,width=0.8\textwidth]{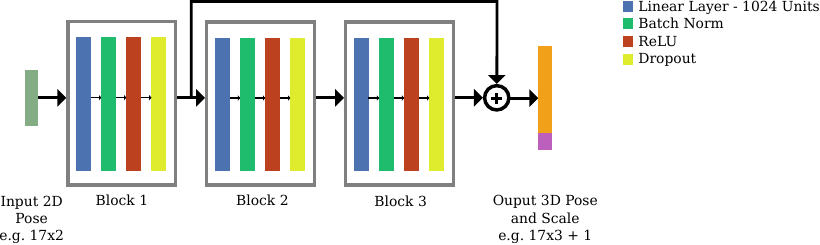}

    \caption{Network architecture. We use a similar architecture to \cite{MartinezAEstimation} but include scale prediction at the end of the network.}
    \label{fig:supp_network}
\end{figure*}

We use the same fully connected network architecture as \cite{MartinezAEstimation} for all experiments.
We use the one stage version of the model (Fig.~\ref{fig:supp_network}) as we only observed a minor loss in performance compared the two stage version. 
To predict the scale parameter $s$ used in our reprojection loss we add an additional fully connected layer to the output of the penultimate set of layers and apply a sigmoid non-linearity to its output. The output of the non-linearity is scaled using a hyperparameter $r$ to allow the network to predict an arbitrarily wide range. For the default method `Ours Relative', detailed in Tab.~1 of the main paper, we set $r=1$.
In the relative depth loss we set $\lambda=2.5$.
We set the weighting hyperparameters $\alpha$ and $\gamma$ in the main loss to 1.0 and set $\beta$ to 0.1. 
We train all models on Human3.6M for 25 epochs, as we observe that they do not tend to benefit from additional training time. We train our relative model from scratch on LSP for 100 epochs.
For our relative model we center the input 2D keypoints by setting the root location to (0,0).
We did not perform this centering for the supervised baseline as we found that it hurt performance, but we did center the 3D coordinates in a similar fashion. 
As in \cite{MartinezAEstimation}, we clip the gradients to 1.0 during training. 
Training time on Human3.6M is less than five minutes for one epoch for our relative model.

\subsection{Human Annotation Performance}
\begin{figure*}[t]
    \centering
    \subfigure[Accuracy vs Distance]{  
        \centering
        \includegraphics[trim={10px 15px 10px 30px},clip,width=.23\textwidth]{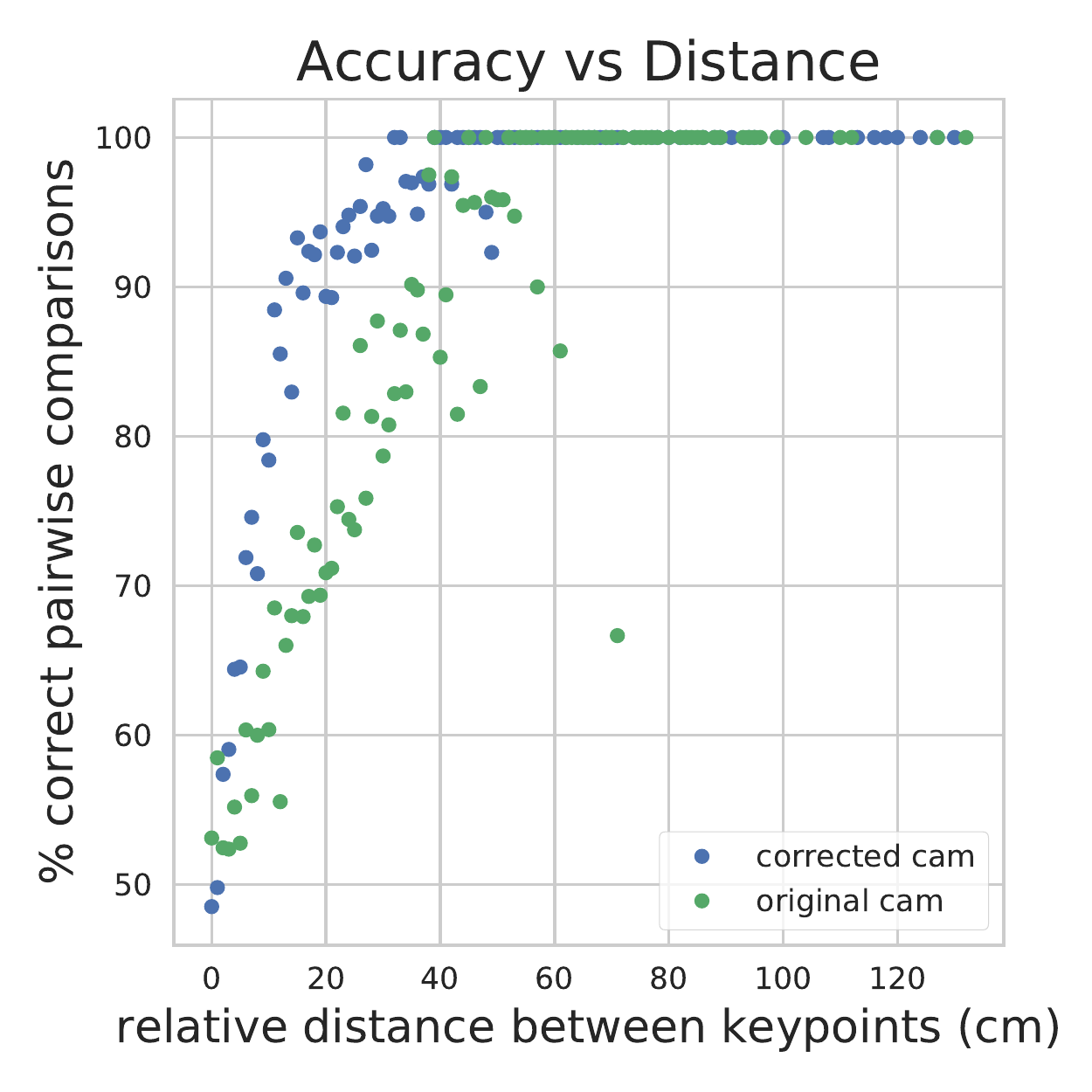}
    }~
    \subfigure[Annotator Accuracy]{
        \centering
        \includegraphics[trim={10px 15px 10px 30px},clip,width=.46\textwidth]{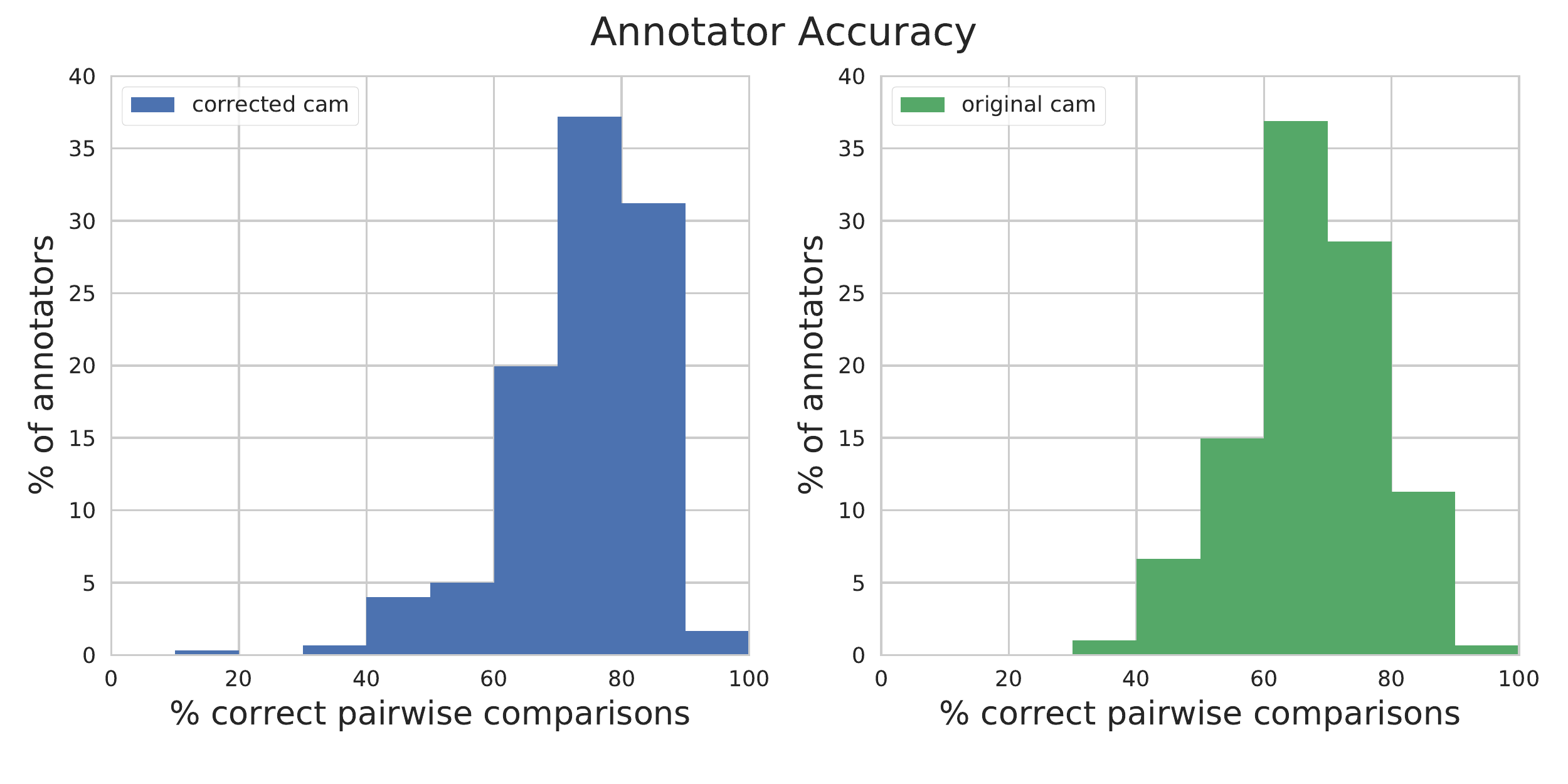}
    }~
    \subfigure[Camera Correction]{
        \centering
        \includegraphics[trim={0px 0px 0px 0px},clip,width=.23\textwidth]{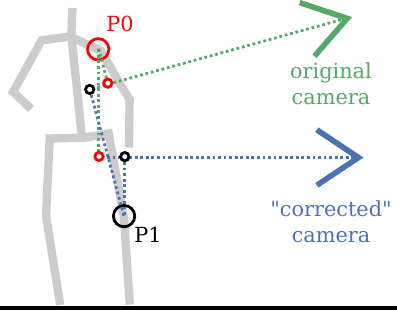}
    }
    \caption{Human relative depth annotation performance on 1,000 images selected from the Human3.6M dataset \cite{ionescu2014human3}. (a-b) Without correcting for the orientation of the camera, annotators perform worse (green lines). (c) The green camera represents the input view and the blue is the upright orientated view as perceived by our annotators. If the camera is orientated upwards when performing the evaluation the relative depths are a better match to the annotator provided labels.}
    \label{fig:supp_user_study}
\end{figure*}

In this section we provide some additional discussion of the results of our user study on Human3.6M~\cite{ionescu2014human3}.
As mentioned in the main paper, we observed that annotators tended to estimate depth in images by correcting for the orientation of the camera. 
In Fig.~\ref{fig:supp_user_study} (c) we see an illustration of this effect. 
Here, from the perspective of the original camera view in green the keypoint `P0' is closer than `P1'. 
In practice, even though they see an image of the scene taken from the perspective of the green camera, annotators seemingly correct for the orientation of the camera and `imagine' the distance of the scene from the perspective of the blue camera. 
While this change in camera position is subtle, it affects the relative ordering of the points as `P1' is now closer to the camera. 
We hypothesize that this is a result of the annotator imagining themselves in the same pose as the individual in the image and then estimating the distance to the camera in a Manhattan world sense. 
Without correcting for this effect 67\% of the provided pairwise annotations are correct, but when this is taken into account then accuracy increases to 71\%. 
We correct for the bias by forcing the camera to be upright when computing the scene depth. 
The results before and after applying this correction and annotator accuracies can be viewed in Figs. \ref{fig:supp_user_study} (a) and (b).
This effect is likely to be exacerbated in Human3.6M as there are only four different camera viewpoints in the entire dataset and they are all facing downwards. 
We expect this to be less of an issue for datasets that feature a larger variation in camera viewpoints relative to the subject of interest as the dominant ground plane will have less of a biasing effect.
% mention numbers in LSP if relevant

\begin{figure*}[h]
    \centering
    \subfigure{  
        \centering
    \includegraphics[trim={90px 20px 90px 20px},clip,width=1.0\textwidth]{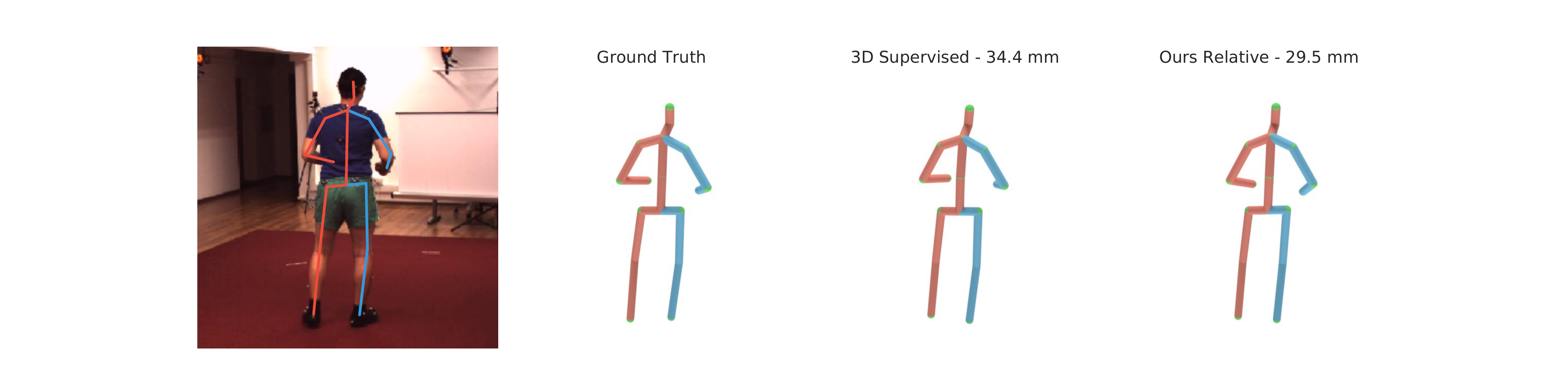}
    }\\\vspace{-10pt}
        \subfigure{  
        \centering
    \includegraphics[trim={90px 20px 90px 20px},clip,width=1.0\textwidth]{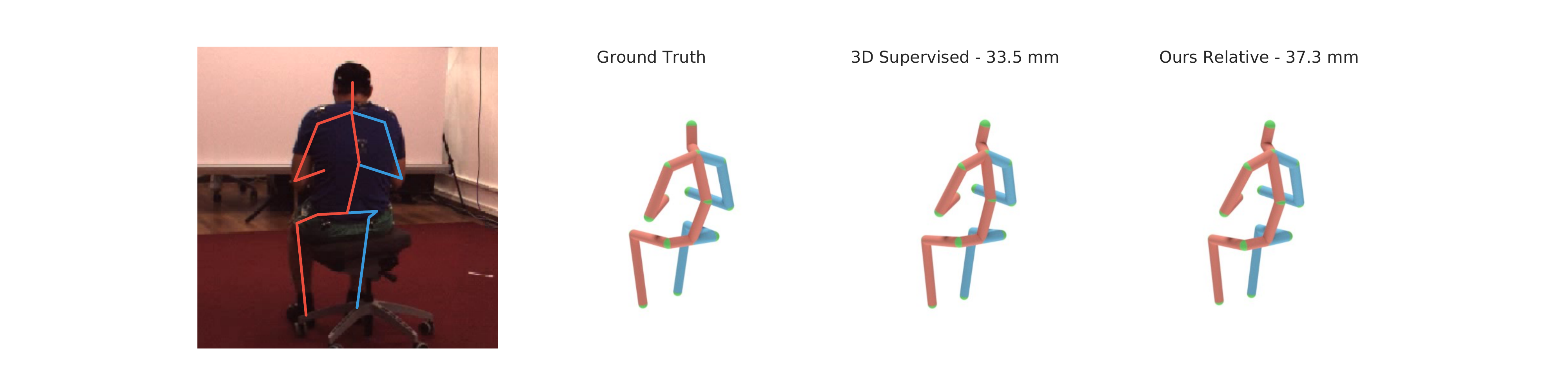}
    }\\\vspace{-10pt}
        \subfigure{  
        \centering
    \includegraphics[trim={90px 20px 90px 20px},clip,width=1.0\textwidth]{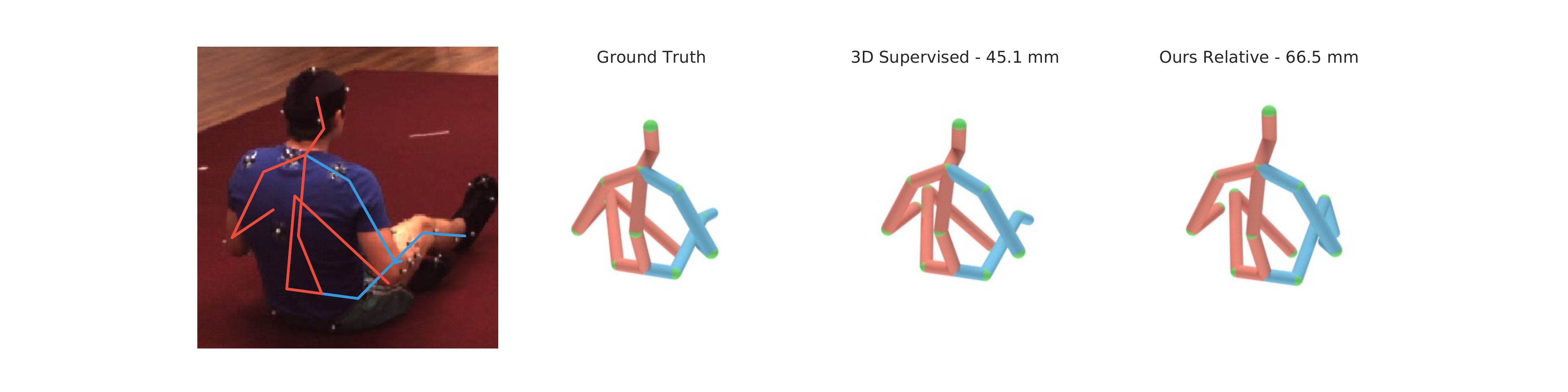}
    }\\\vspace{-10pt}
        \subfigure{  
        \centering
    \includegraphics[trim={90px 20px 90px 20px},clip,width=1.0\textwidth]{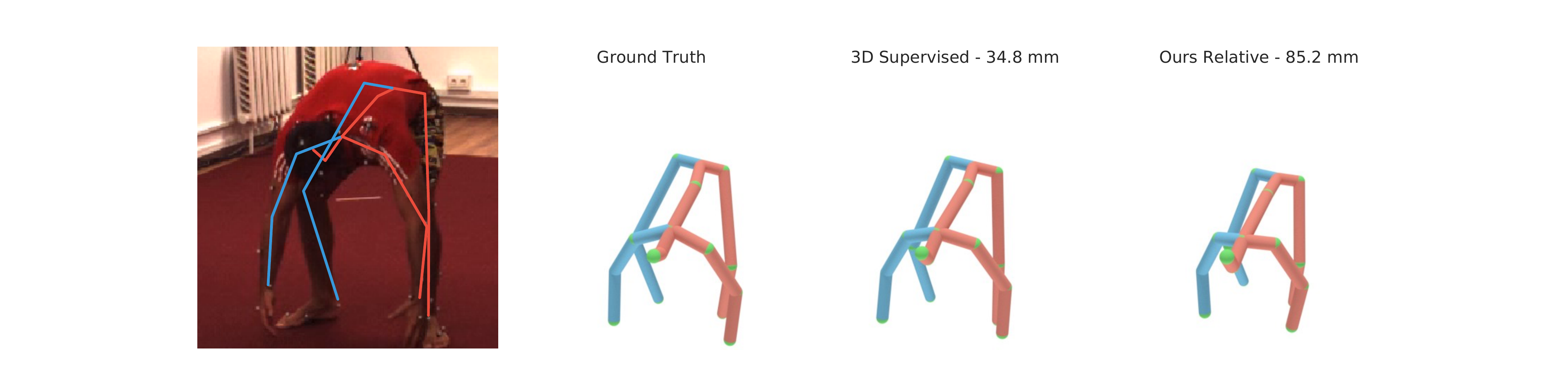}
    }\\\vspace{-10pt}
    \subfigure{
        \centering
    \includegraphics[trim={90px 20px 90px 20px},clip,width=1.0\textwidth]{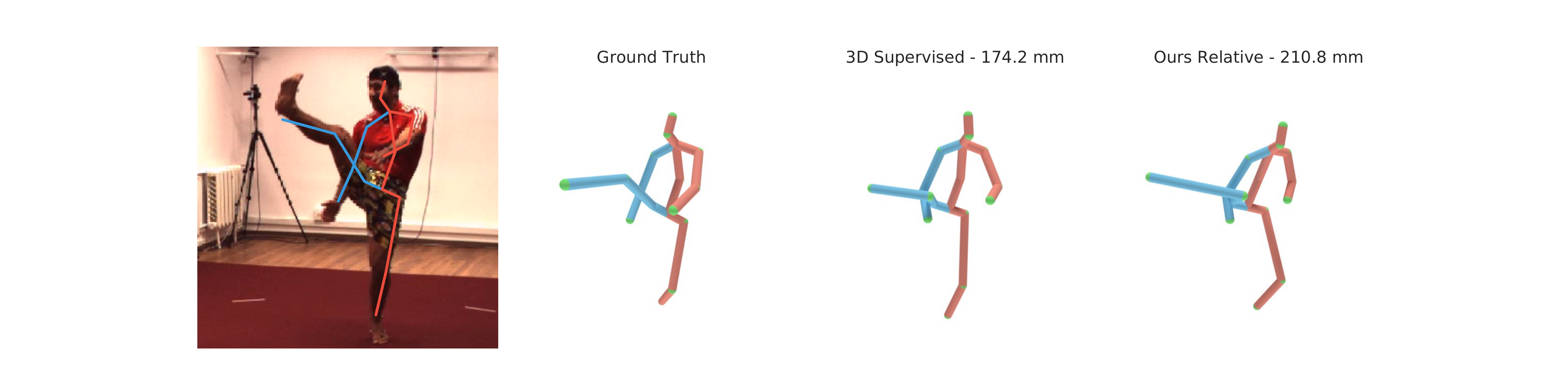}
    }
    \caption{Test time predictions on Human3.6M. Despite using much weaker training data our relative model (Ours Relative 17j GT/GT) produces sensible results for most input poses. Both the supervised and our approach are depicted after rigid alignment, with the pose error displayed on top.}
    \label{fig:supp_h36_results}
\end{figure*}

\begin{figure*}[h]
    \centering
    \subfigure[Easy keypoint pairs.]{  
        \centering
    \includegraphics[trim={100px 30px 90px 5px},clip,width=1.0\textwidth]{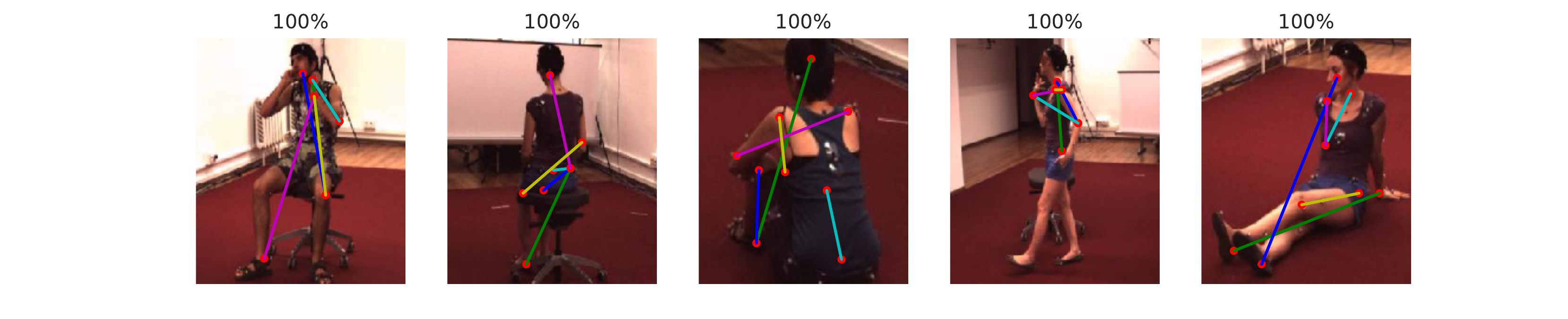}
    }\\\vspace{-10pt}
    \subfigure[Difficult keypoint pairs.]{
        \centering
    \includegraphics[trim={100px 30px 90px 5px},clip,width=1.0\textwidth]{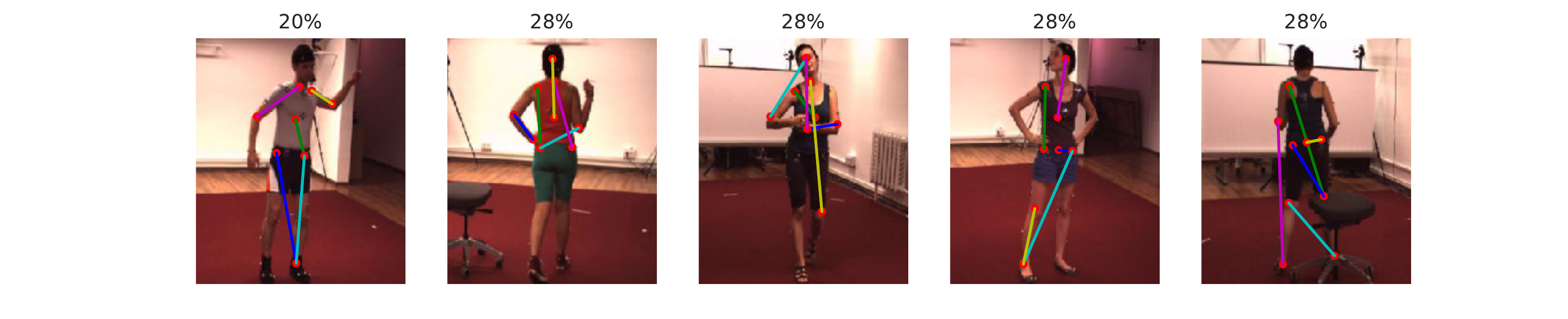}
    }
    \caption{Example Human3.6M keypoint pairs. Colored lines link pairs that were shown to annotators.
    The numbers on top represent the raw accuracy of the crowd provided labels before merging.
    (a) Easy pairs where no incorrect annotations were made i.e. each of the five annotators annotated all five pairs correctly.
    (b) Difficult examples where the randomly selected pairs tend to be at a similar distance to the camera, resulting in lower performance.}
    \label{fig:supp_h36_easy_diff}
\end{figure*}

\begin{figure*}[h]
    \centering
    \includegraphics[trim={0px 0px 0px 0px},clip,width=0.85\textwidth]{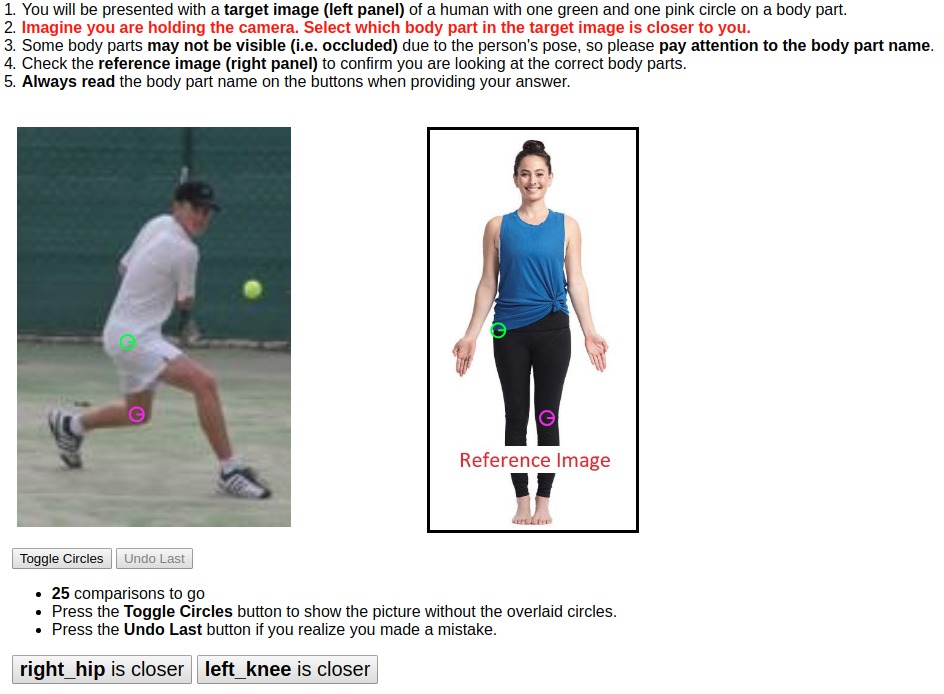}

    \caption{Our user interface. The annotator's goal is the determine the relative depth of the highlighted keypoints in the left image. The reference image on the right highlights the same keypoints and helps in situations where they are occluded.}
    \label{fig:supp_turk_interface}
\end{figure*}

Fig.~\ref{fig:supp_turk_interface} depicts an example task from our user interface that was shown to annotators. 
The first time annotators performed our task they were presented with a short tutorial that included sample images and were instructed on how to use the interface and given feedback when they predicted the incorrect depth ordering.
For each task, we also included a short delay before annotators could select their response to encourage them to pay attention to the input image when performing the task.
Example annotations from Human3.6M \cite{ionescu2014human3} can be seen in Fig.~\ref{fig:supp_h36_easy_diff}. 
Unsurprisingly, keypoint pairs that have larger relative distances are easier to annotation.
For these examples the ground truth accuracies are computed with respect to the corrected ground truth. 
Example 3D predicted poses on Human3.6M can be seen in Fig.~\ref{fig:supp_h36_results}.

% 4.993 average annotations per image
% 5 median annotations per image

% 301 workers
% 82.940 average annotations per worker
% 25 median annotations per worker

% 24965 annotations
% 12301 annotations == 1
% 12664 annotations == 0
% 3629 / 5000 (72.58%) images are finished
% Corrected 0.4988 % pos GT out of 5000
% Original 0.4896 % pos GT out of 5000
% overall num annotations       24965
% numer of images               1000
% num of pairs                  5000
% acc (overall)                 0.7142
% acc merged  (per pair)        0.7894
% acc merged_orig (per pair)    0.701
% acc mv      (per pair)        0.781

 % trim={<left> <lower> <right> <upper>}

%%%%%%%%%%%%%%%%%%%%%%%%%%%%%%%%%%%%%%%%%%%%%%%%%
% LSP QUALITATIVE
% <left> <lower> <right> <upper>
\begin{figure*}[t]
    \centering
    \subfigure{  
        \centering
        \includegraphics[trim={140px 30px 85px 40px},clip,width=.45\textwidth]{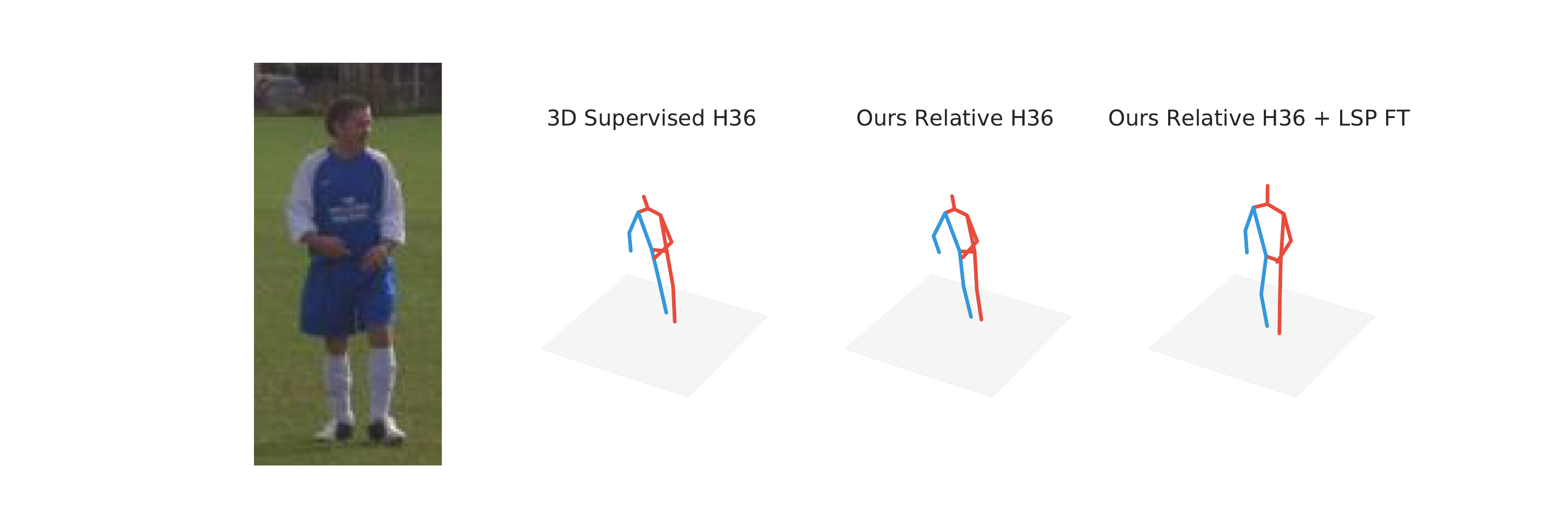}
    }~
    \subfigure{
        \centering
        \includegraphics[trim={140px 30px 85px 40px},clip,width=.45\textwidth]{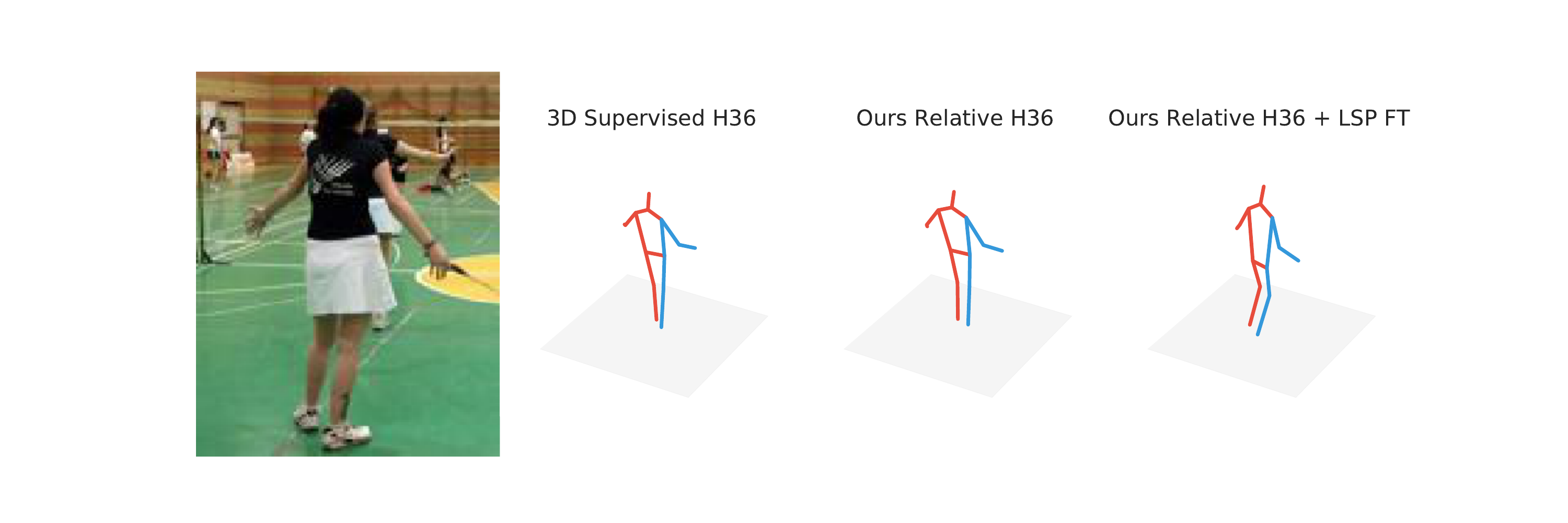}
    }\\
    \subfigure{  
        \centering
        \includegraphics[trim={140px 30px 85px 40px},clip,width=.45\textwidth]{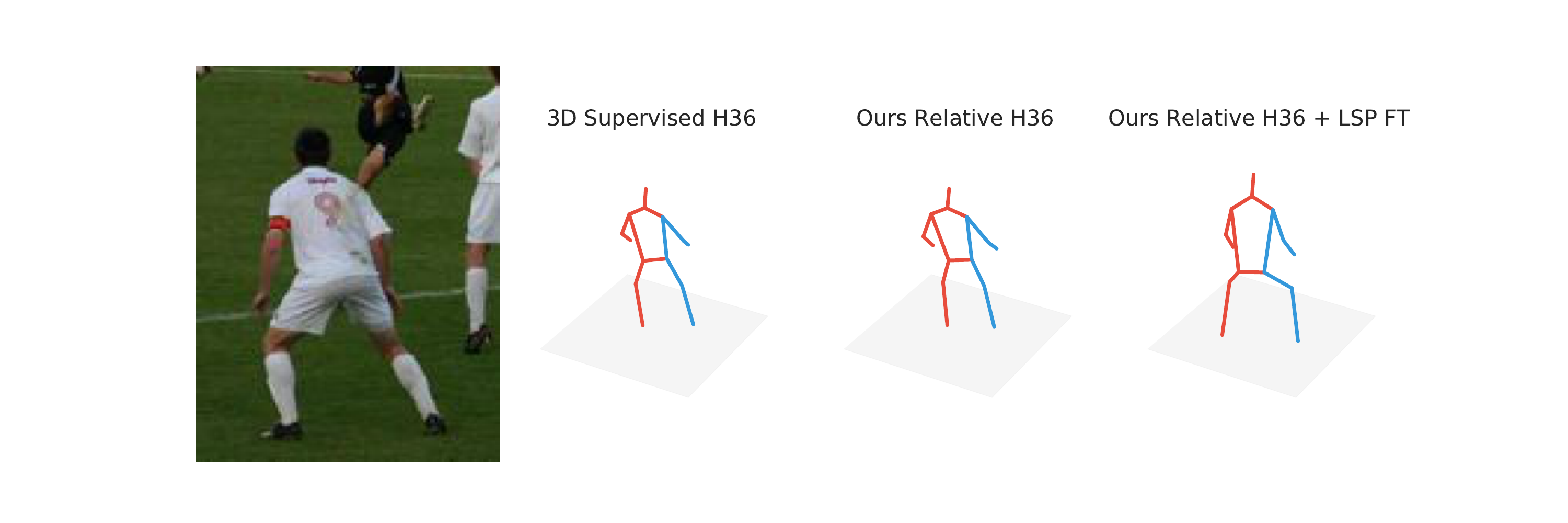}
    }~
    \subfigure{
        \centering
        \includegraphics[trim={140px 30px 85px 40px},clip,width=.45\textwidth]{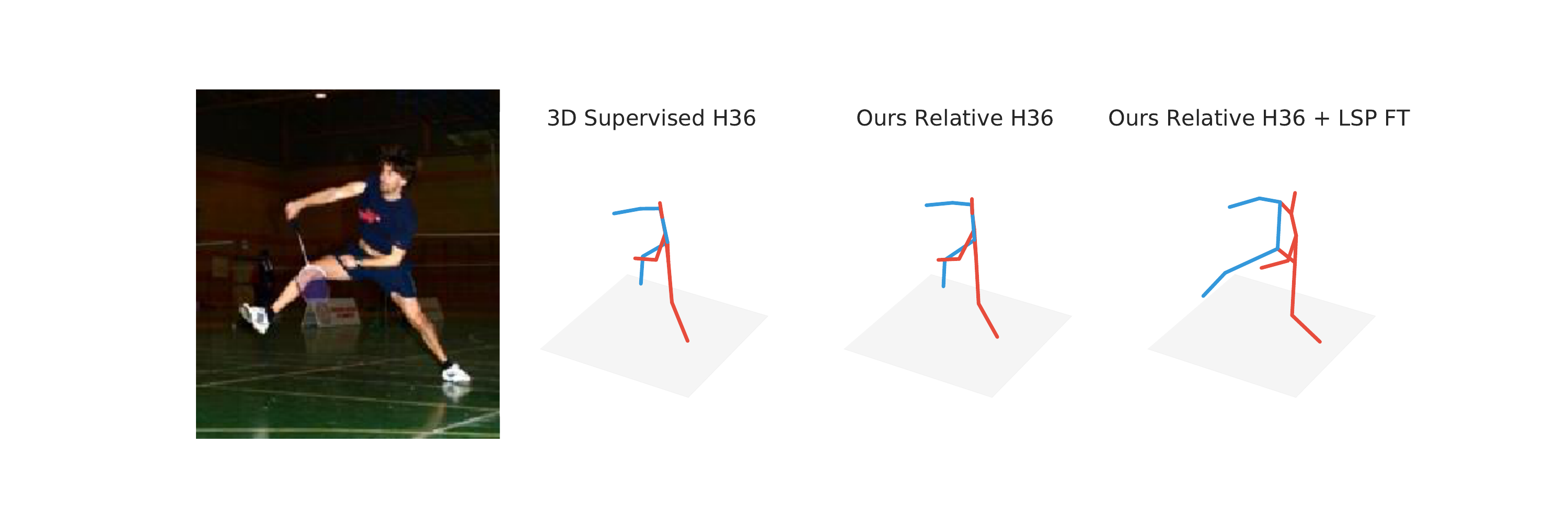}
    }\\
     \subfigure{
       \centering
        \includegraphics[trim={140px 30px 85px 40px},clip,width=.45\textwidth]{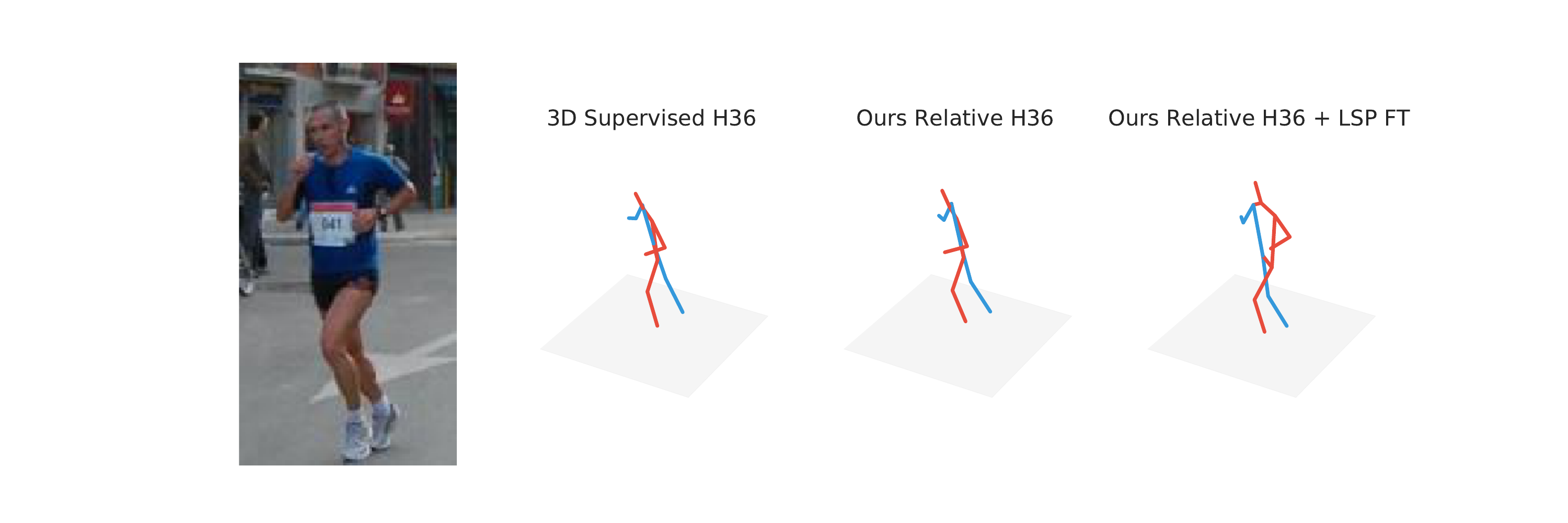}
    }~
    \subfigure{
        \centering
        \includegraphics[trim={140px 30px 85px 40px},clip,width=.45\textwidth]{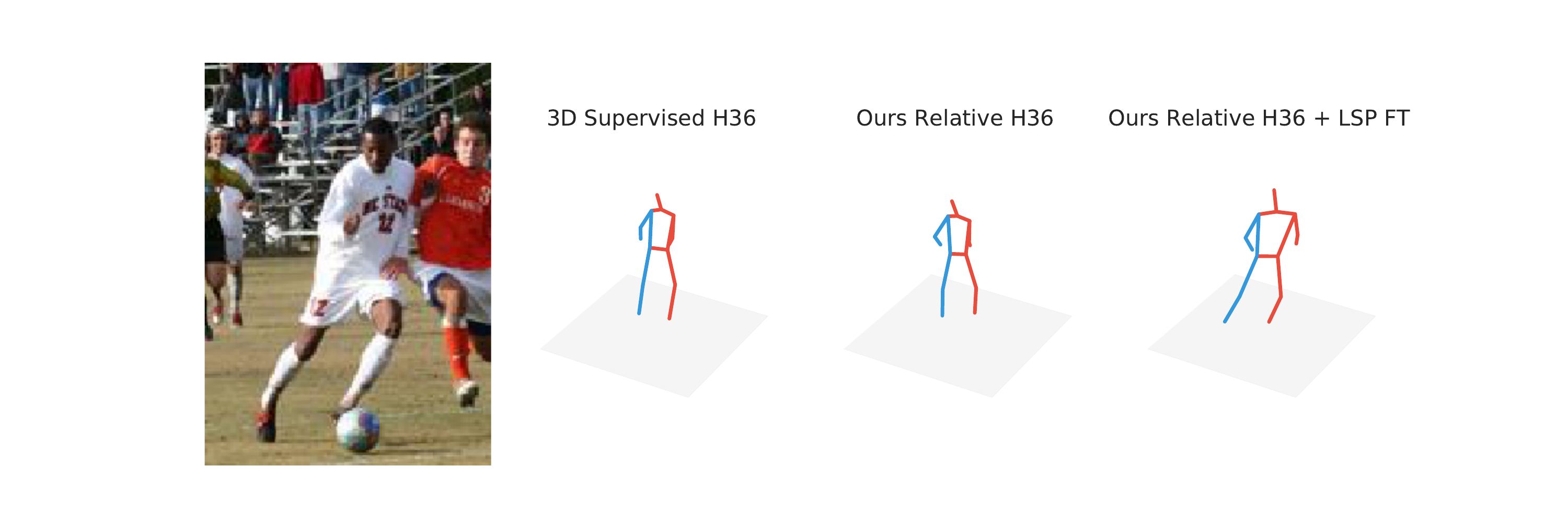}
    }
    \subfigure{  
        \centering
        \includegraphics[trim={140px 30px 85px 40px},clip,width=.45\textwidth]{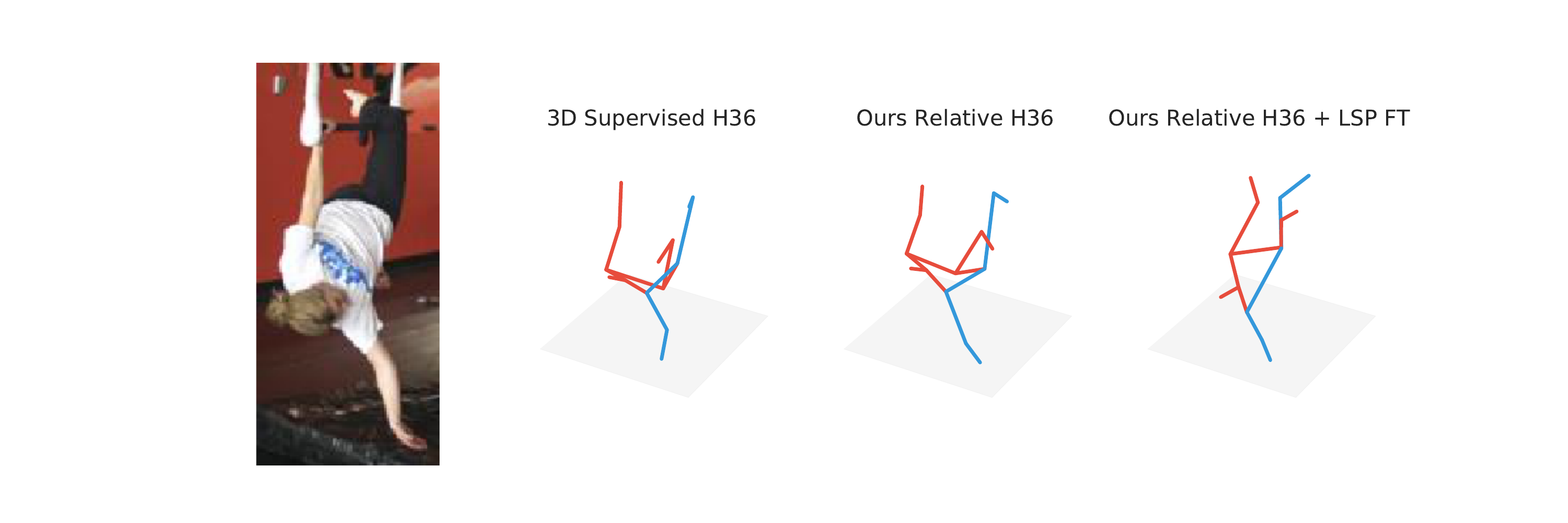}
    }~
    \subfigure{
        \centering
        \includegraphics[trim={140px 30px 85px 40px},clip,width=.45\textwidth]{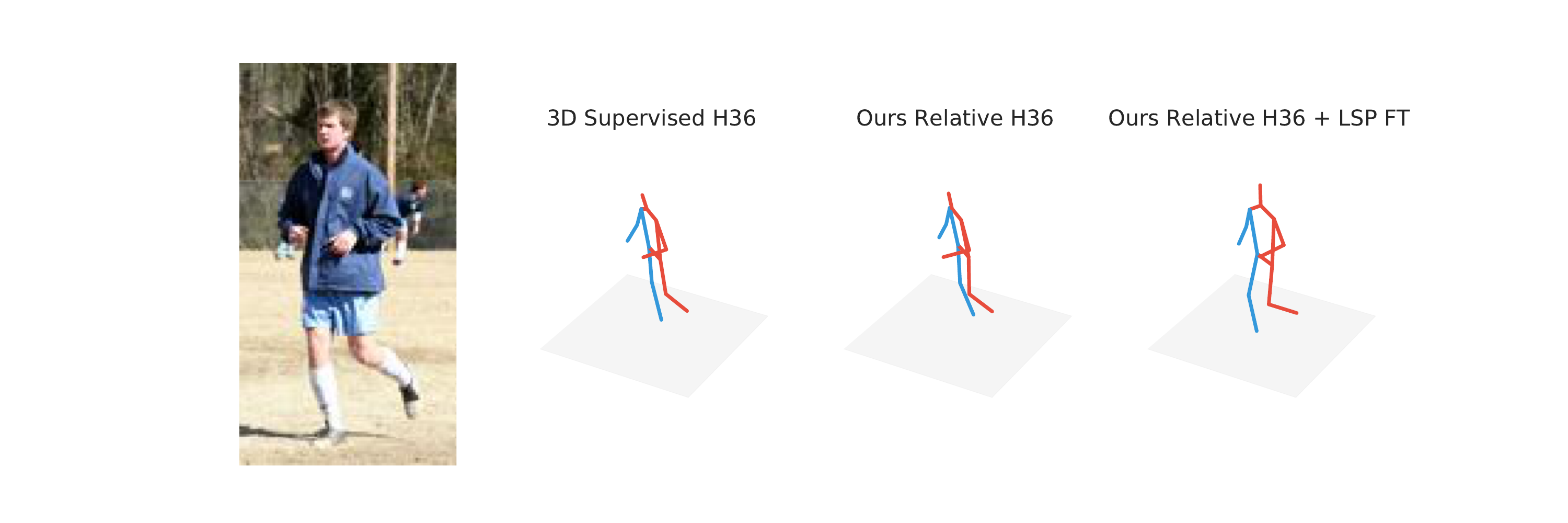}
    }\\
    \subfigure{  
        \centering
        \includegraphics[trim={140px 30px 85px 40px},clip,width=.45\textwidth]{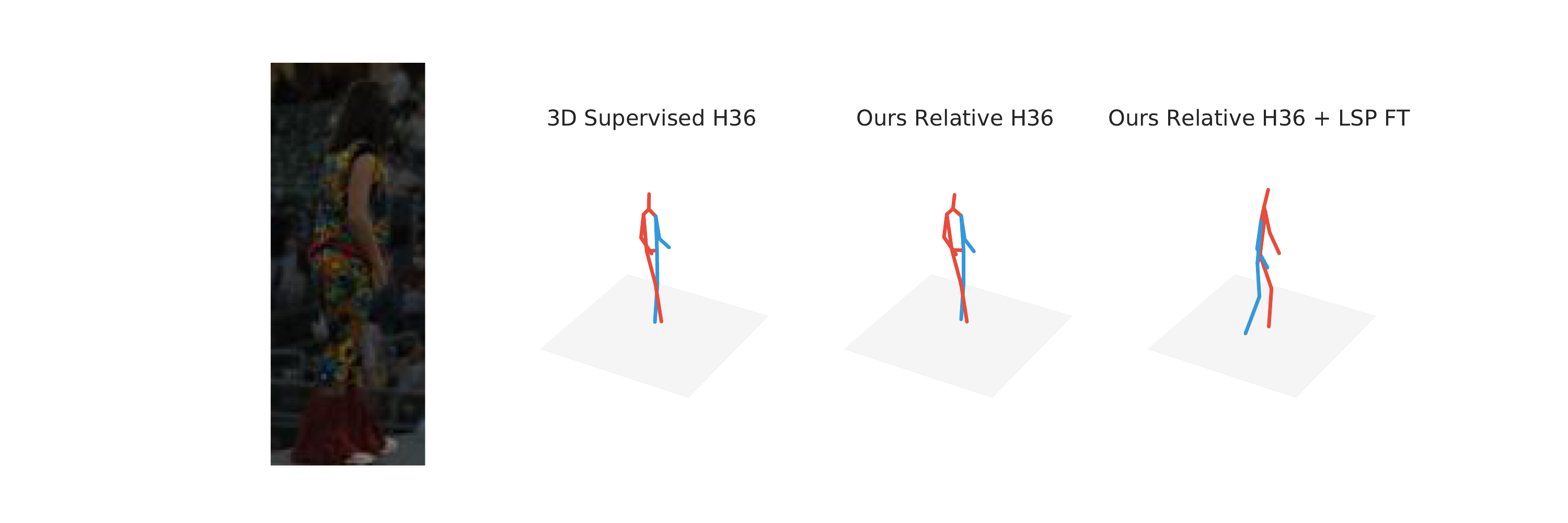}
    }~
    \subfigure{
        \centering
        \includegraphics[trim={140px 30px 85px 40px},clip,width=.45\textwidth]{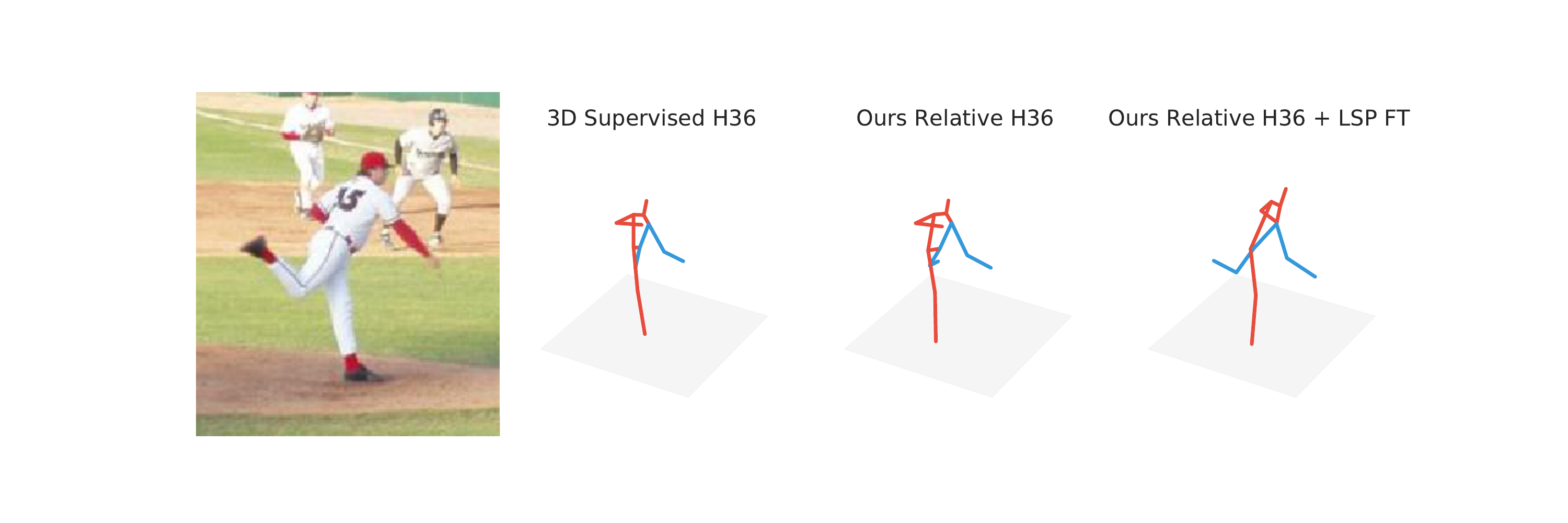}
    }\\
     \subfigure{
       \centering
        \includegraphics[trim={140px 30px 85px 40px},clip,width=.45\textwidth]{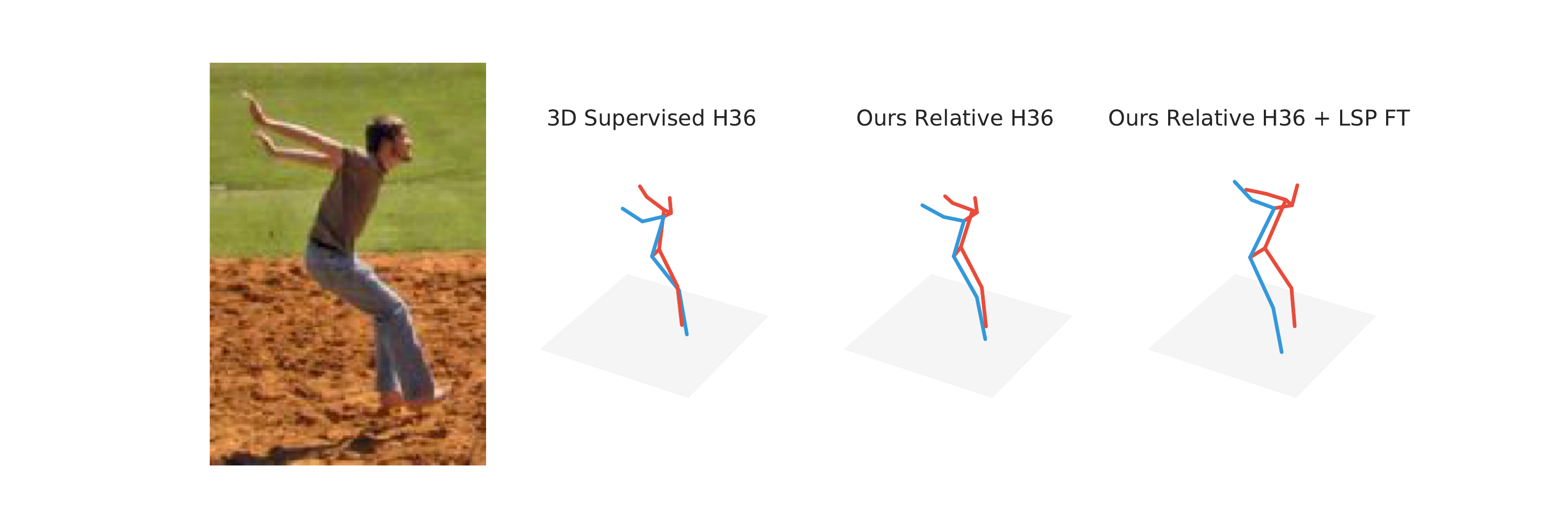}
    }~
    \subfigure{
        \centering
        \includegraphics[trim={140px 30px 85px 40px},clip,width=.45\textwidth]{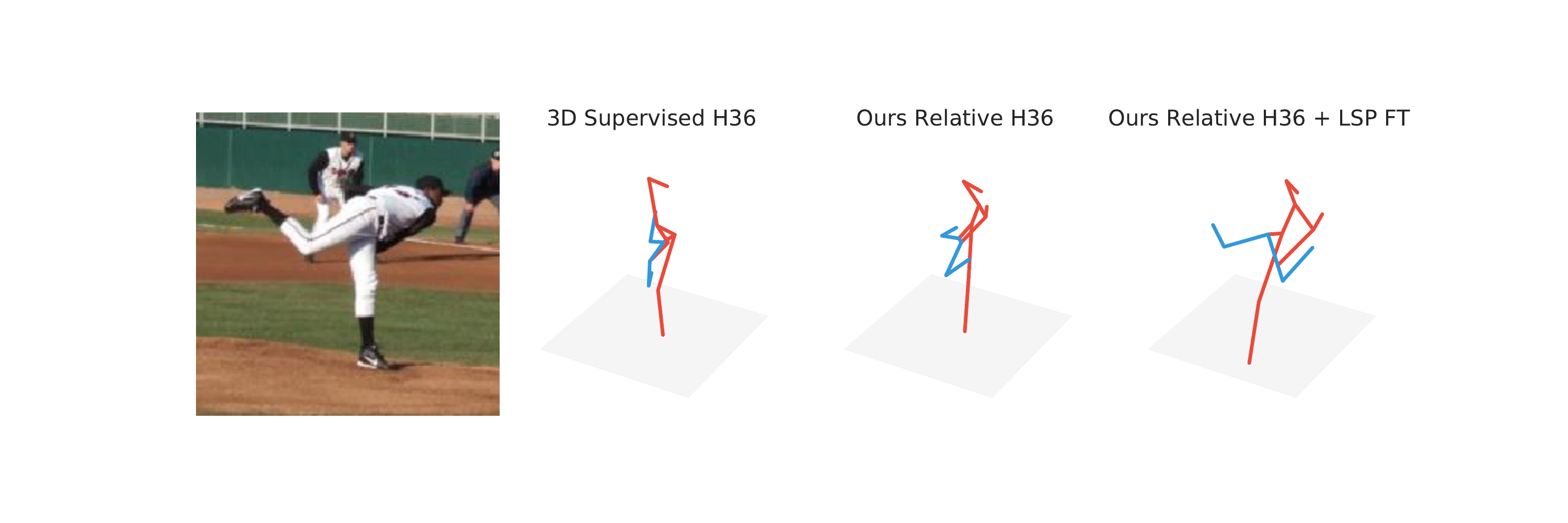}
    }\\
     \subfigure{
       \centering
        \includegraphics[trim={140px 30px 85px 40px},clip,width=.45\textwidth]{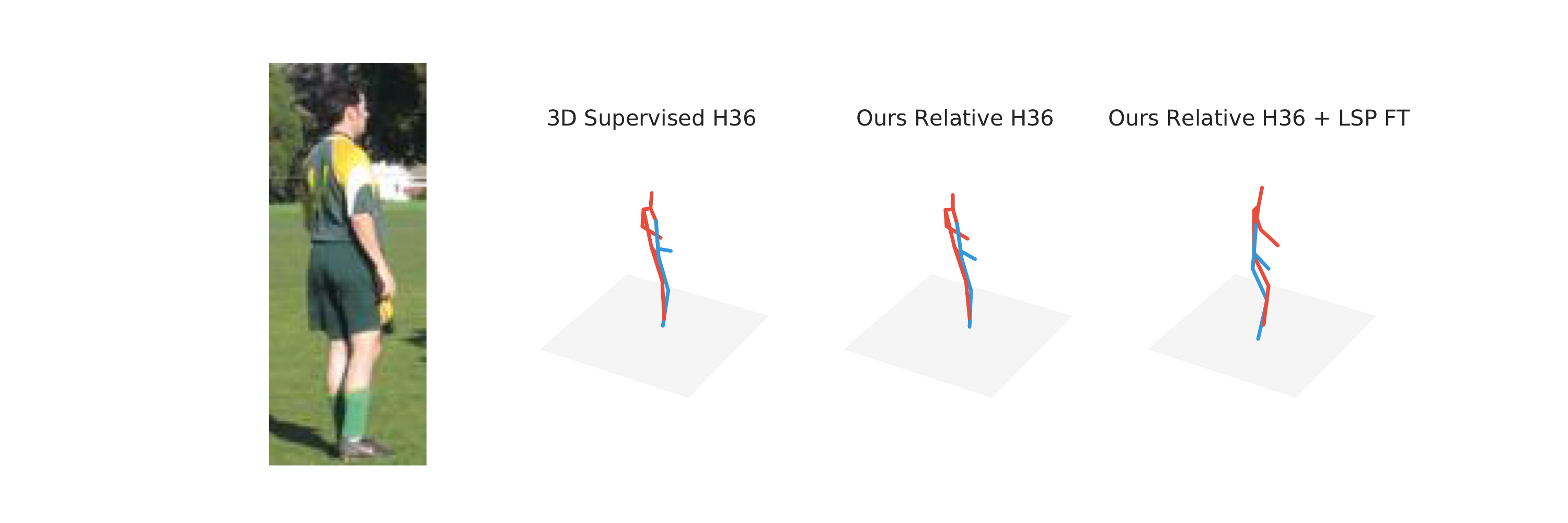}
    }~
    \subfigure{
        \centering
        \includegraphics[trim={140px 30px 85px 40px},clip,width=.45\textwidth]{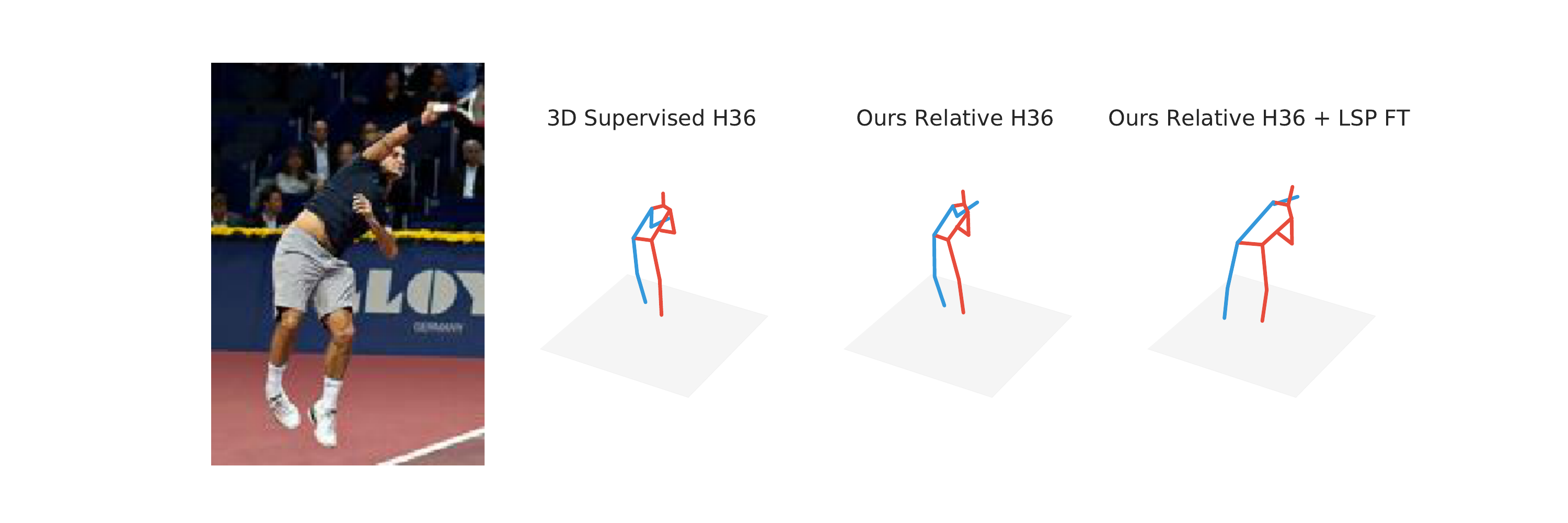}
    }

\caption{Predicted 3D poses on LSP. Fine-tuning (FT) on LSP significantly improves the quality of our predictions especially for images containing uncommon poses and viewpoints that are not found in Human3.6M.}
\end{figure*}

\end{document}